\documentclass[journal]{IEEEtran}

\usepackage{lscape}

\usepackage{multirow}     %
\usepackage{booktabs}     %
\usepackage{amsmath,amssymb,amsfonts}
\usepackage{rotating}
\usepackage{graphicx}
\usepackage{algorithm}
\usepackage{algorithmic}
\usepackage{adjustbox}
\usepackage{tabularray}
\usepackage[graphicx]{realboxes}
\usepackage{diagbox, subcaption}
\usepackage{xcolor}
\usepackage{array}
\newcolumntype{H}{>{\setbox0=\hbox\bgroup}c<{\egroup}@{}}

\begin{document}

\bstctlcite{IEEEexample:BSTcontrol}
\title{ASC: Adaptive Scale Feature Map Compression For Deep Neural Network
}

\author{\IEEEauthorblockN{Yuan Yao, and Tian-Sheuan Chang, \textit{Senior Member, IEEE}}
\thanks{This work was supported by the National Science and Technology Council, Taiwan, under Grant 111-2622-8-A49-018-SB, 110-2221-E-A49-148-MY3, and 110-2218-E-A49-015-MBK. The authors are affiliated with the Institute of Electronics, National Yang Ming Chiao Tung University, Taiwan. (e-mail: yuan.yao.ee10@nycu.edu.tw, tschang@nycu.edu.tw) }%
\thanks{To be published in IEEE Transactions on Circuits and Systems I: Regular Papers, doi 10.1109/TCSI.2023.3337283}
}

\maketitle

\begin{abstract}%

Deep-learning accelerators are increasingly in demand; however, their performance is constrained by the size of the feature map, leading to high bandwidth requirements and large buffer sizes. We propose an adaptive scale feature map compression technique leveraging the unique properties of the feature map. This technique adopts independent channel indexing given the weak channel correlation and utilizes a cubical-like block shape to benefit from strong local correlations. The method further optimizes compression using a switchable endpoint mode and adaptive scale interpolation to handle unimodal data distributions, both with and without outliers. This results in 4$\times$ and up to 7.69$\times$ compression rates for 16-bit data in constant and variable bitrates, respectively. Our hardware design minimizes area cost by adjusting interpolation scales, which facilitates hardware sharing among interpolation points. Additionally, we introduce a threshold concept for straightforward interpolation, preventing the need for intricate hardware. The TSMC 28nm implementation showcases an equivalent gate count of 6135 for the 8-bit version. Furthermore, the hardware architecture scales effectively, with only a sublinear increase in area cost. Achieving a 32$\times$ throughput increase meets the theoretical bandwidth of DDR5-6400 at just 7.65$\times$ the hardware cost.

~ 
\end{abstract}

\section{Introduction}
Deep learning accelerators (DLAs) enable wide adoption of AI to various applications. However, its massive amount of feature maps demands high memory bandwidth, which becomes the performance bottleneck for real time execution, especially for edge devices. As shown in Table~\ref{comparison_FM}, the amount of feature maps will become larger and larger for higher resolution input and complex tasks. For example, the real time execution of EDSR will need 504GB/s DRAM bandwidth for 30 frames per second and 16-bit data width. 
A promising approach is to compress feature maps on-the-fly during processing.

\begin{table}[htbp]
\centering
\caption{Number of the feature map and parameters in various tasks. Feature maps are counted twice, as they are both stored and fetched, while parameters are only fetched. The actual bitrate shall multiply these numbers with bitwidth and frames per second.}
\begin{adjustbox}{width=\columnwidth,center}
\begin{tabular}{c|c|c|c|cH}
\hline
\textbf{Task} &
  \textbf{Network} &
  \textbf{Input   size} &
  \textbf{\begin{tabular}[c]{@{}c@{}}\#Weights\end{tabular}} &
  \textbf{\begin{tabular}[c]{@{}c@{}}\#Feature\end{tabular}} &
  \textbf{\begin{tabular}[c]{@{}c@{}}Ratio\\ (F/W)\end{tabular}} \\ \hline
\multirow{4}{*}{Classification} & AlexNet   & \multirow{4}{*}{224$\times$224} & 62.3M & 0.66M    & 0.021x  \\
                                & VGG16     &                          & 138M  & 13.57M   & 0.196x  \\
                                & ResNet34  &                          & 21.3M & 2.96M    & 0.277x  \\
                                & ViT-small &                          & 21.6M & 16.34M   & 0.756x  \\ \hline
\begin{tabular}[c]{@{}c@{}}Semantic\\ Segmentation\end{tabular} &
  \begin{tabular}[c]{@{}c@{}}DeepLabv3+   \\ (ResNet50)\end{tabular} &
  513$\times$513 &
  39.7M &
  150.82M &
  3.798x \\ \hline
Super   Resolution              & EDSR      & 1024$\times$768                 & 43M   & 8493.46M & 395.04x \\ \hline
\end{tabular}
\label{comparison_FM}
\end{adjustbox}
\end{table}

Different feature map compression methods have been proposed: the lossless method, and the lossy method in either the time or frequency domain. Lossless methods such as EBPC~\cite{EBPC} or AEC~\cite{Area_efficient} leverage statistical redundancy to represent data in a reversible manner, ensuring that no information is lost during the compression process. However, their compression rates highly depend on input sparsity, which is only true for ReLU-based networks. Scaling up EBPC~\cite{EBPC} and AEC~\cite{Area_efficient} can be costly, exhibiting quadratic complexity due to the presence of a variable symbol size compressor.  Additionally, AEC~\cite{Area_efficient} is only applicable to the floating-point format. 

However, lossy methods sacrifice non-essential details to achieve size reduction, leading to a trade-off between information preservation and size reduction. These methods can be executed in the time domain, as seen with Least-Squares Fitting Compression (LSFC)~\cite{least-square-fitting} or revised texture compression~\cite{real-time-superresolution}. Alternatively, they can be applied in the frequency domain with different transforms~\cite{Wavelet-compression, DCT-memoryefficient}. However, LSFC offers a limited compression rate.  The study in \cite{real-time-superresolution} modified S3TC~\cite{S3TC_patent} to have 32-point interpolation, but also leads to a low compression rate. The approach in \cite{Wavelet-compression} only supports 1x1 convolutions. Moreover, the transforms presented in ~\cite{Wavelet-compression, DCT-memoryefficient} come with a significant computational cost.

Existing work has several weaknesses: limited hardware scalability, variable compression rates, and limited support for various types of layers and number formats. To address these issues, it is imperative to identify a compression and decompression scheme that demonstrates excellent hardware scalability. Additionally, a fixed compression rate is crucial for seamless integration into DLA. Finally, the chosen scheme should offer enhanced support for a diverse range of layer types and formats.

This paper proposes an adaptive scale feature map compression (ASC) based on S3TC, but tailored to feature map characteristics to address memory bandwidth and buffer size bottlenecks. The proposed method employs independent channel indexing due to weaker channel correlation and a cubical-like block shape to enhance strong local correlations, further augmented by similarity-based reordering. It introduces a switchable endpoint mode for dynamic sparse layers and adaptive scale interpolation for unimodal distributions, both with and without outliers. Compatible with both fixed-point and floating-point data types, the technique delivers compression ratios of up to 4$\times$ in constant bitrate and 7.69$\times$ in variable bitrate, achieving nearly lossless performance. Despite the simplicity of this algorithm, we further reduce the required hardware cost significantly using revised linear scale and shifted scale methods. The design, implemented using TSMC 28nm technology, necessitates only 6135 gate counts and achieves a 32$\times$ throughput increase with just 7.65$\times$ the hardware cost.

The remainder of the paper is organized as follows. Section II presents the proposed algorithm. Section III details the corresponding hardware design. Section IV presents the experimental results. Finally, Section V concludes the paper.

\section{Proposed Method}
\subsection{Review of S3TC}

\begin{figure}[htbp]
\centering
\includegraphics[height=!,width=1.0\linewidth,keepaspectratio=true]{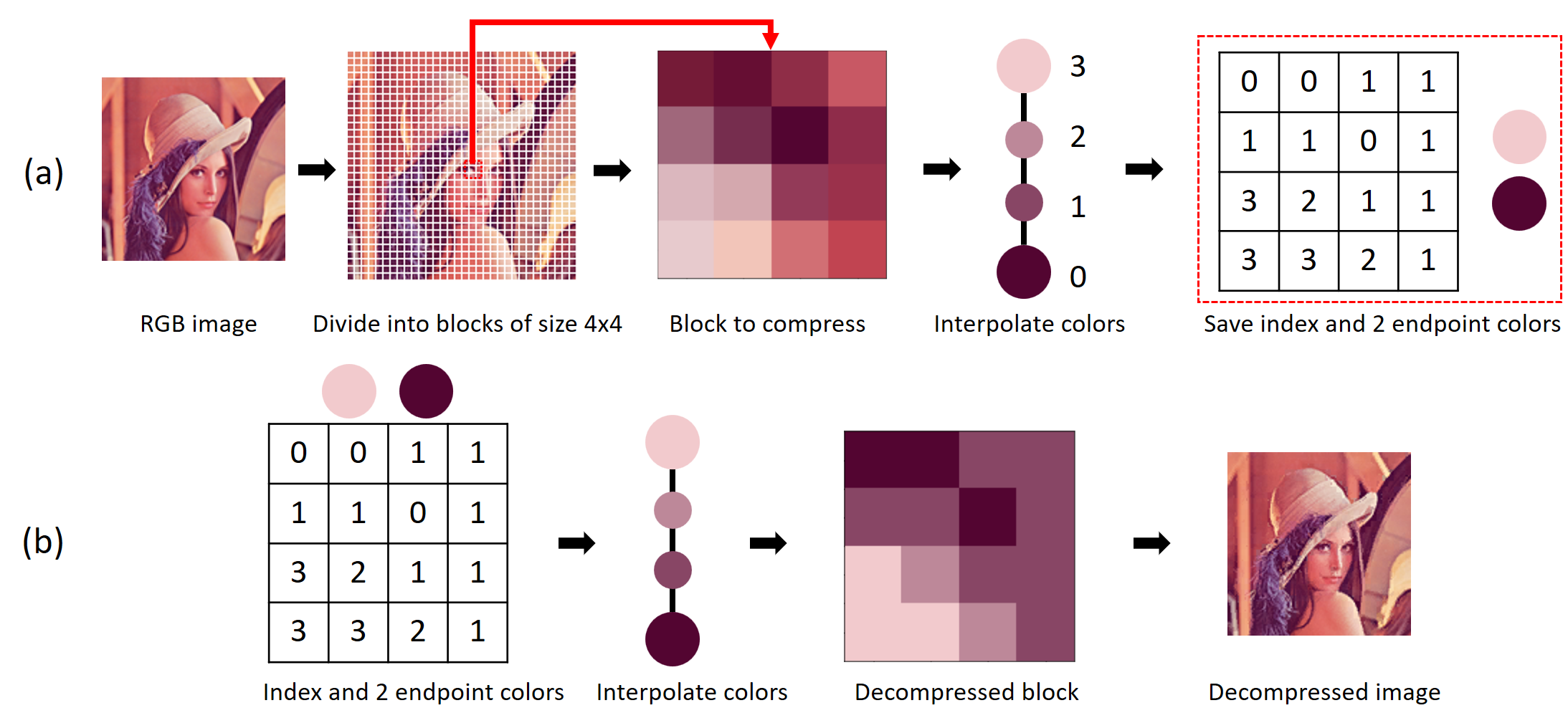}
\caption{(a) S3TC encoding process, (b) S3TC decoding process}
\label{S3TC_detail}
\end{figure}

Fig.\ref{S3TC_detail}(a) illustrates the S3TC encoding process. The input image, in the RGB888 format, is partitioned into 4$\times$4 blocks. Within each block, the maximum and minimum values for the three channels are identified. These values form two endpoints: one from the combined maximums and another from the minimums. For improved compression, the endpoints are truncated to the RGB565 format. Two intermediate colors are then interpolated between these endpoints, yielding four distinct colors, each represented by a two-bit index. Pixels are mapped to the nearest color index. The final encoding comprises the two endpoints and the sixteen indices. The decoding process, depicted in Fig.\ref{S3TC_detail}(b), is more straightforward. Intermediate colors are derived from the stored endpoints. Each index is substituted with its corresponding color, resulting in the decompressed image once all blocks are processed.

\subsection{Challenges and Proposed Solutions}
While S3TC offers a consistent compression rate, a simple algorithm, and regular data access patterns, it is specifically tailored for RGB images. Feature maps differ in several aspects, such as channel count, channel correlation, sparsity, and smoothness, which hinder the efficient direct application of S3TC. To address these challenges, we introduce ASC, which we detail in the subsequent sections.

\textbf{Channel count:} While images typically consist of three channels (RGB), feature maps can vary in channel count depending on the models and layers. For instance, deeper layers often possess more channels. To accommodate this variability and enhance performance, we suggest adopting a \textit{cubical-like block shape} for optimization. This approach improves the efficiency of compressing feature maps across diverse models and layers.

\textbf{Channel Correlation:} RGB channels in images exhibit strong interchannel correlations, allowing values in one channel to predict those in others~\cite{channelcorrelation}. As a result, the S3TC index is shared across channels, achieving a higher compression rate. On the contrary, feature maps tend to have weaker channel correlations. To avoid performance degradation, we advocate for \textit{independent channel indexing} over S3TC's shared indexing approach. Furthermore, we introduce a \textit{similarity-based reordering} algorithm to further enhance performance.

\textbf{Sparsity Level:} Owing to the widespread use of the Rectified Linear Unit (ReLU) activation function in deep learning, the negative values in the feature maps are truncated, resulting in sparsity, a characteristic absent in the images. We introduce a \textit{switchable endpoint mode} that allows users to decide whether to leverage this sparsity based on their model requirements.

\textbf{Smoothness:} Natural images often exhibit smooth transitions between pixel values. On the contrary, certain feature map layers might not present such smoothness, containing a higher number of outliers. To address this disparity, we propose the \textit{adaptive scale interpolation} method, tailored for both smooth and non-smooth feature maps. This technique ensures accurate compression and decompression, taking into account the unique characteristics of each type of feature map.

\subsection{Overview of ASC}
In our approach, we introduce two versions: constant bitrate (ASC-CBR) and variable bitrate (ASC-VBR). ASC-CBR ensures a consistent compression rate and boasts superior hardware scalability. On the other hand, ASC-VBR, which integrates ASC-CBR with zero value compression (ZVC), leverages feature map sparsity to achieve a higher compression rate. However, this increased rate comes at the expense of a fixed compression rate and hardware scalability. In this subsection, we primarily focus on ASC-CBR, while ASC-VBR will be detailed in a subsequent subsection.

\begin{figure*}[htbp]
\centering
\includegraphics[trim={0cm 3cm 1cm 0cm},clip, height=!,width=1.0\linewidth,keepaspectratio=true]{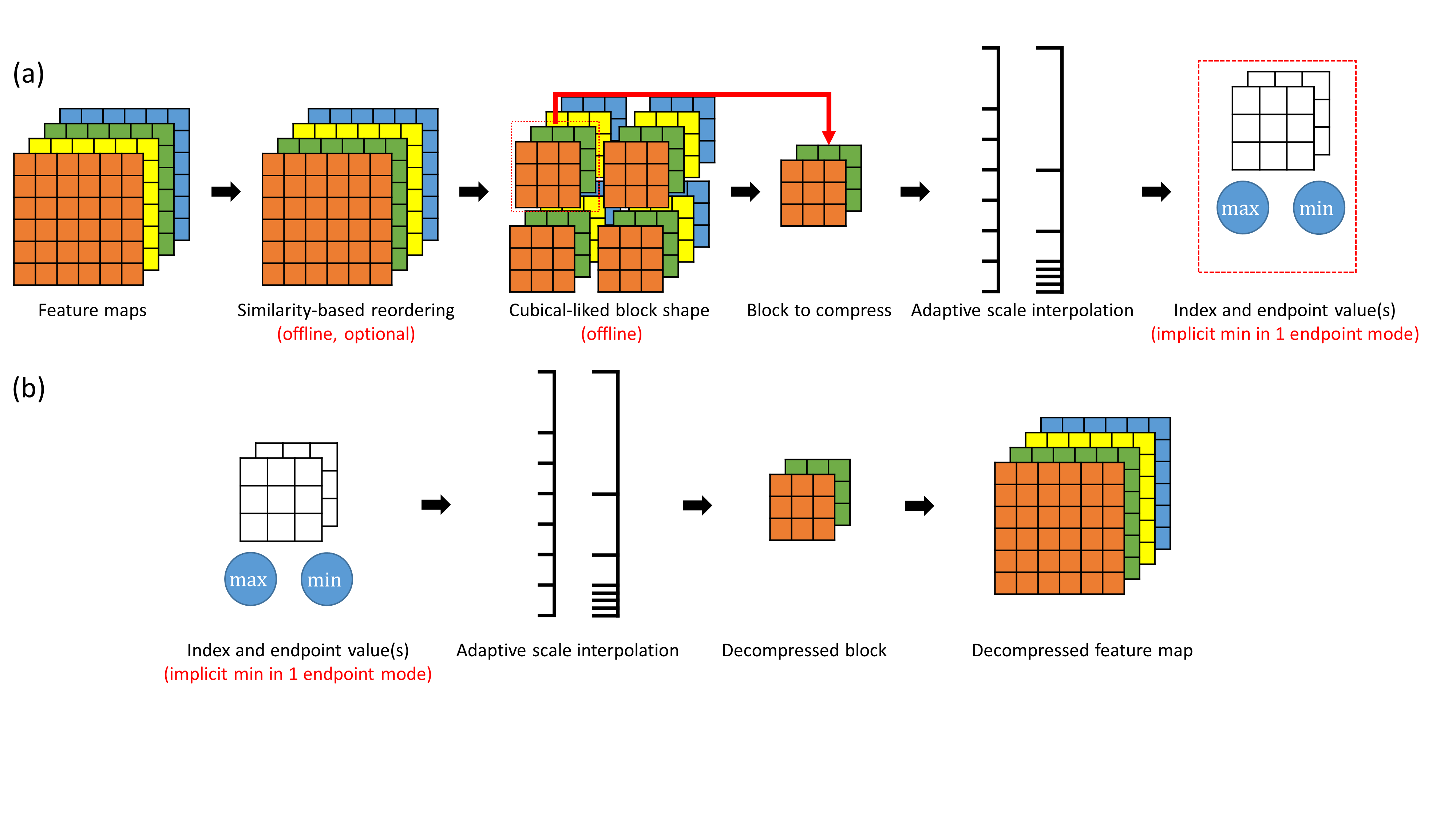}
\caption{Proposed ASC-CBR processes: (a) encoding and (b) decoding}
\label{ASC_overview}
\end{figure*}

Fig.~\ref{ASC_overview} delineates the ASC-CBR algorithm. Before encoding, an optional \textit{similarity-based reordering} is employed to enhance channel correlation in adjacent feature maps. During encoding, the feature map is segmented into blocks according to the \textit{cubical-like block shape} recommendation. Endpoint determination is based on the selected endpoint mode. Subsequently, \textit{adaptive scale interpolation} formulates six intermediate values. Pixels are then assigned to their nearest value index, with the scale determined by the L1 error between the compressed and original data. Notably, value indices are channel-specific and are not shared. The process concludes with the storage of endpoints and indices. During decoding, the six intermediate values are interpolated using the stored endpoints and the chosen scale. Each index is then substituted with its corresponding value.

Subsequent sections delve into the specific techniques integrated into the algorithm. The experimental results in this section adhere to the approach outlined in Section IV. For simplicity, we will use VGG16 and ResNet34 in the following results for their popularity. All models used in this section are quantized to INT8 for both activation and weight.

\subsection{Channel Indexing and Reordering}
\subsubsection{Independent Channel Indexing}
\begin{figure}[htbp]
\centering
\includegraphics[trim={4cm 0 4cm 0},height=!,width=0.7\linewidth,keepaspectratio=true]{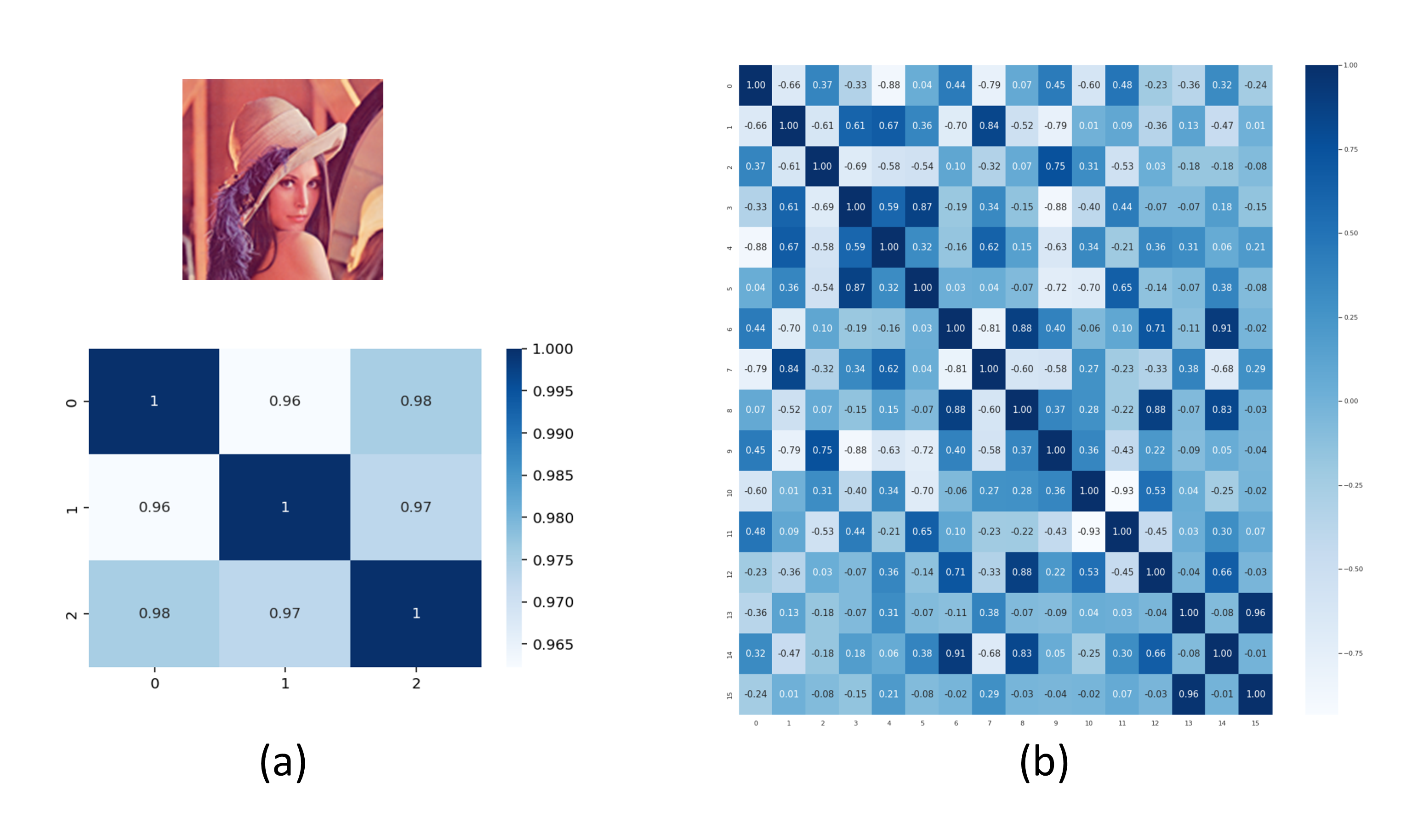}
\caption{(a) Similarity matrix for an image, (b) Similarity matrix for a feature map}
\label{similarity_difference}
\end{figure}

RGB images exhibit strong channel correlations, implying that values in one channel can often predict those in others~\cite{channelcorrelation}. Such correlations justify the use of a single index to compress the three channels. However, the efficacy of this technique for feature maps depends on their degree of channel correlation. To determine its suitability, a thorough examination of these correlations is essential. Fig.~\ref{similarity_difference} shows similarity matrices for both image and feature maps. While pronounced similarities are evident among the RGB channels of the image, the feature map shows a notably diminished inter-channel similarity.

Based on the observations, we transition from S3TC's shared channel indexing to \textit{independent channel indexing}. This approach assigns unique indices to values at identical positions across different channels, ensuring minimal performance loss while preserving algorithmic efficacy.

\subsubsection{Similarity-based Reordering}
Despite the diminished correlation among feature map channels, significant correlation persists among certain channels. By grouping these channels together, their values tend to be similar, aiding interpolation with minimal loss, as demonstrated in our \textit{cubical-like block shape} discussed later. However, the arrangement of channels in a model is typically arbitrary. To facilitate more effective grouping, we introduce \textit{similarity-based reordering}.

\textit{Similarity-based reordering} arranges channels based on their similarity. To identify the best-matched pairs, we explore two methods: greedy and heuristic. The greedy method selects the maximum element from the similarity matrix in each iteration, representing the most correlated channel pair. After a pair is chosen, the corresponding rows and columns are removed to avoid repetition. This procedure continues until no channels remain in the matrix. However, this approach might neglect channels that have multiple strong correlations, potentially leading to suboptimal results.

\begin{figure}[htbp]
\centering
\includegraphics[height=!,width=1.0\linewidth,keepaspectratio=true]{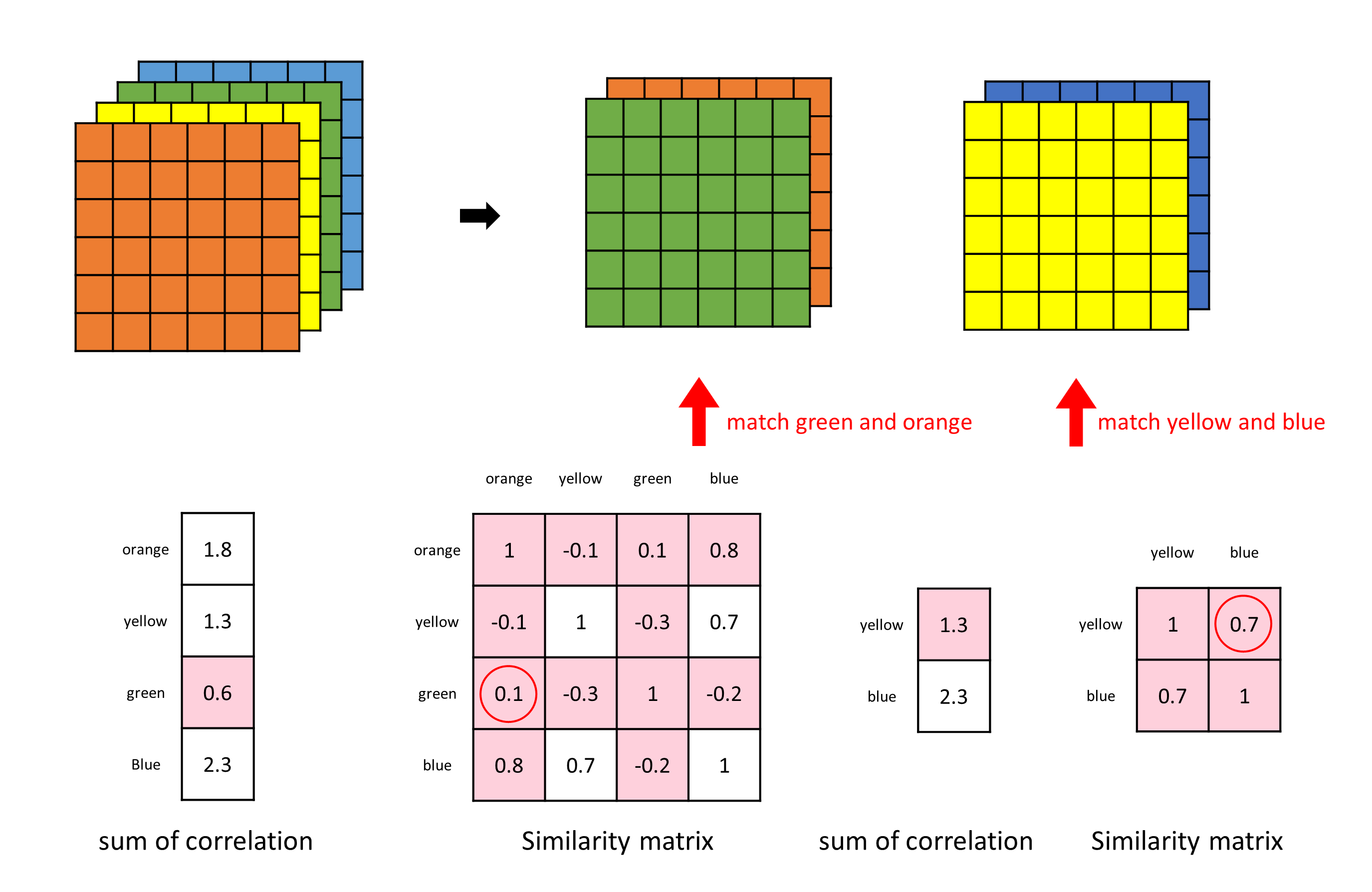}
\caption{The heuristic method to match two channels}
\label{reordering_heuristic}
\end{figure}

Instead of merely selecting the largest element, the heuristic method calculates the sum of correlations for each channel and uses this aggregate to guide the selection process, as illustrated in Fig.~\ref{reordering_heuristic}. Channels with a higher cumulative correlation suggest strong affiliations with other channels, implying that pairing them in later iterations would not harm performance. In contrast, channels with a lower sum of correlations are perceived as highly isolated. We prioritize pairing these channels early to ensure that their limited connections aren't preemptively selected by other channels.

In our example, the green channel, which has the lowest correlation sum, is prioritized for pairing. The most significant element in its row is 0.1, indicating a modest correlation with the orange channel. In the subsequent iteration, the most prominent element for the yellow channel, 0.7, signifies a strong correlation with the blue channel. This approach exemplifies our strategy: by prioritizing the pairing of isolated channels, we improve match quality, thereby enhancing the overall efficiency of reordering. Each matched pair forms a group. By applying this method iteratively, we optimize the kernel sequence. It's worth noting that this reordering is optional and is executed offline after model training.

\subsection{Cubical-Like Block Shape}
As illustrated in Fig.\ref{similarity_difference}, the degree of inter-channel correlation is contingent upon specific channels. This correlation becomes even more pronounced within a smaller region\cite{least-square-fitting}. Such observations suggest that grouping neighboring pixels can reveal a higher degree of redundancy that can then be eliminated. To capitalize on these correlations, our algorithm aims to make the block shape as cubical as possible.

We characterize the block shape using the notation (blockW, blockH, blockC), where blockW denotes the width, blockH signifies the height, and blockC indicates the number of channels in a block. The block$_{size}$ is defined as the total number of values within a block, computed as block$_{size}$ = blockW$\times$blockH$\times$blockC. Given that the number of channels in feature maps is predominantly a power of two, we assume block$_{size}$ to also be a power of two in the subsequent examples.

\subsection{Switchable Endpoint Mode}

Given the diverse levels of sparsity, we introduce the \textit{switchable endpoint mode}. For models with high sparsity, we utilize the one-endpoint mode, where only the maximum endpoint is calculated and stored, while the minimum endpoint is assumed to be zero. This approach yields a higher compression rate with minimal performance degradation. Conversely, for dense models, we adopt the two-endpoint mode, in which both the maximum and minimum endpoints are calculated and stored. The maximum and minimum endpoints are the maximum and minimum values of the block, respectively. This method more effectively preserves the accuracy of the model. The \textit{switchable endpoint mode} provides users with the flexibility to capitalize on sparsity based on the unique characteristics of their models.

\subsection{Adaptive Scale Interpolation}
\begin{figure}[htbp]
\centering
\includegraphics[height=!,width=1.0\linewidth,keepaspectratio=true]{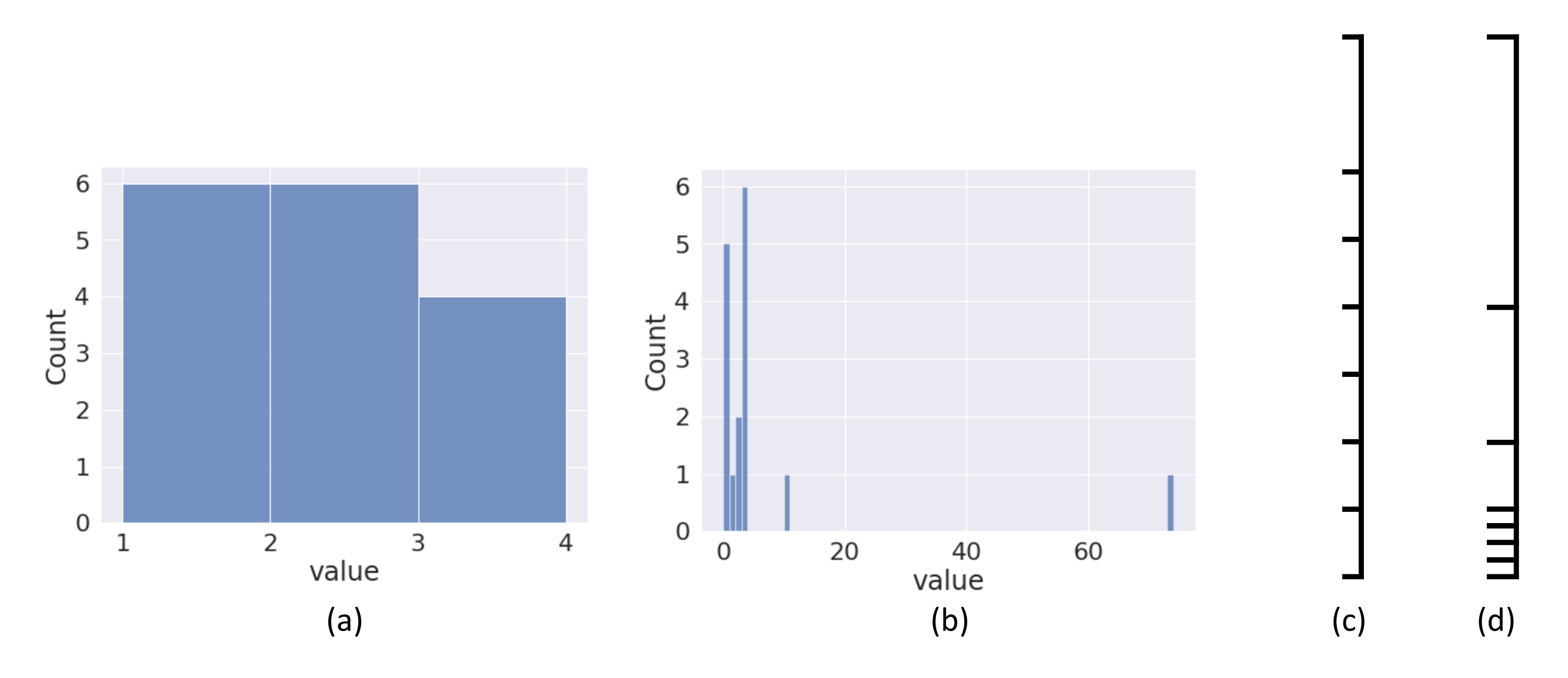}
\caption{(a) Smooth and image-like block, (b) Disjointed block with outliers, (c) Revised linear scale, (d) Log-linear scale}
\label{adaptive_scale}
\end{figure}

Feature maps evolve into semantic information in deeper layers. In the initial layers, the feature map closely mirrors the original image. However, as we progress deeper into the layers, the representation of the object transforms into more abstract feature information, leading to significant deviations from the original image. The early layers exhibit smooth, image-like characteristics with a unimodal distribution confined to a narrow range. In contrast, deeper layers display more fragmented patterns within a broader unimodal distribution, punctuated by outliers. Fig.\ref{adaptive_scale}(a) and (b) depict these two contrasting block types, respectively. Consequently, employing linear interpolation in S3TC for the latter would prove to be less than ideal.

We introduce the \textit{adaptive scale interpolation}, which utilizes two distinct scales for interpolation: a modified linear scale and a log-linear scale. Fig.\ref{adaptive_scale}(c) illustrates the modified linear scale. Unlike the traditional linear scale, where interpolation points are equidistant and suitable for smooth blocks with narrower value ranges, this updated version features a denominator that is a power of two, simplifying hardware implementation. Fig.\ref{adaptive_scale}(d) presents the log-linear scale, tailored to manage outliers by logarithmically decaying interpolated values. Yet, its lower section adopts linear interpolation to cater to the smoother values present in feature maps.

During the compression process, both scales are dynamically applied to blocks, with the selection based on which scale yields a lower L1 loss compared to the original data. This \textit{adaptive scale interpolation} accommodates the distinct characteristics of feature map blocks, ensuring optimal performance for both image-like and non-image-like feature maps.

\subsection{Variable bitrate version: ASC-VBR}
\begin{figure}[htbp]
\centering
\includegraphics[height=!,width=0.9\linewidth,keepaspectratio=true]{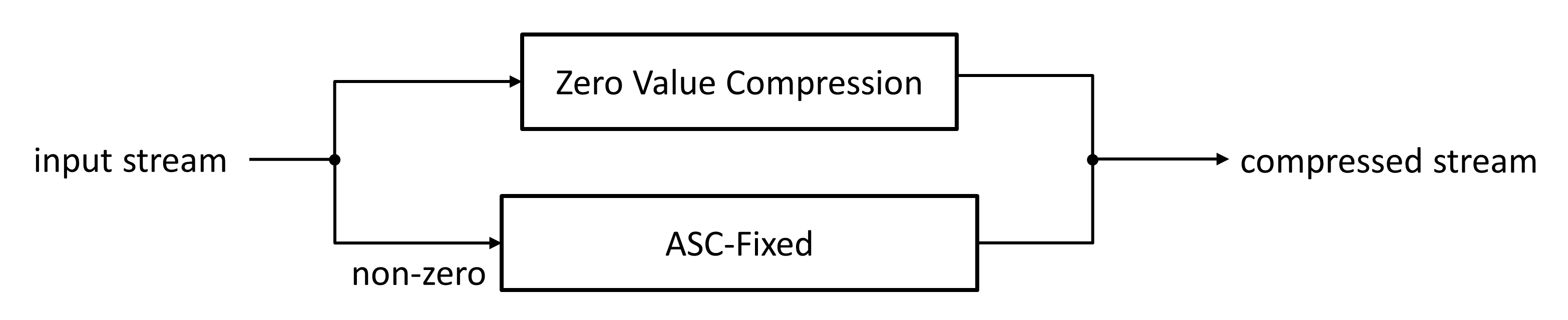}
\caption{Compression scheme of ASC-VBR}
\label{ASC-type2}
\end{figure}

Fig.\ref{ASC-type2} illustrates the variable bitrate version of ASC, termed ASC-VBR. Drawing inspiration from~\cite{EBPC, Area_efficient}, ASC-VBR employs different compression methods for zero and nonzero values. Specifically, zero values undergo compression using zero value compression (ZVC)~\cite{ZVC}, while non-zero values are compressed via ASC-CBR. ZVC leverages bit masking to denote zero and nonzero values, optimizing the bitrate. While ASC-VBR achieves a higher compression rate compared to ASC-CBR, it comes with the trade-offs of diminished hardware scalability and a fluctuating compression rate. Moreover, its efficacy is closely tied to the sparsity level.

\section{Hardware Implementation}
\label{chapter:Hardware}

\subsection{Challenges and Solutions}
Though the proposed ASC is simple, implementing the algorithm in a cost-effective manner presents challenges, particularly in generating interpolation points and identifying the closest interpolation point for a given input. 

A direct approach to the original linear scale interpolation would necessitate the use of dividers, which are notoriously intricate and pose challenges for hardware implementation. To address this, we modify the linear scale to obviate the need for these dividers. However, even with this modification, the formulas for interpolation points within both the revised linear scale and the log-linear scale still demand a plethora of multipliers and adders, providing limited avenues for hardware sharing. To mitigate this challenge, we introduce scale shifting. The interpolation points of the shifted scales bear marked similarities, resulting in a substantial increase in hardware sharing opportunities.

Another challenge emerges when trying to identify the closest interpolation point for a given input. The straightforward method calculates the distances between the input and the interpolation points, subsequently assigning the input to the interpolation point with the shortest distance. This method, however, requires an extensive number of adders, multiplexers, and comparators, leading to exorbitant hardware costs. In contrast, our approach compares the input with pre-established thresholds and then employs a priority encoder to determine the result. This strategy significantly reduces hardware complexity and further supports hardware sharing. The specifics of the proposed design are described below.

\subsection{Proposed ASC Hardware}

\begin{figure}[htbp]
\centering
\includegraphics[height=!,width=1.0\linewidth,keepaspectratio=true]{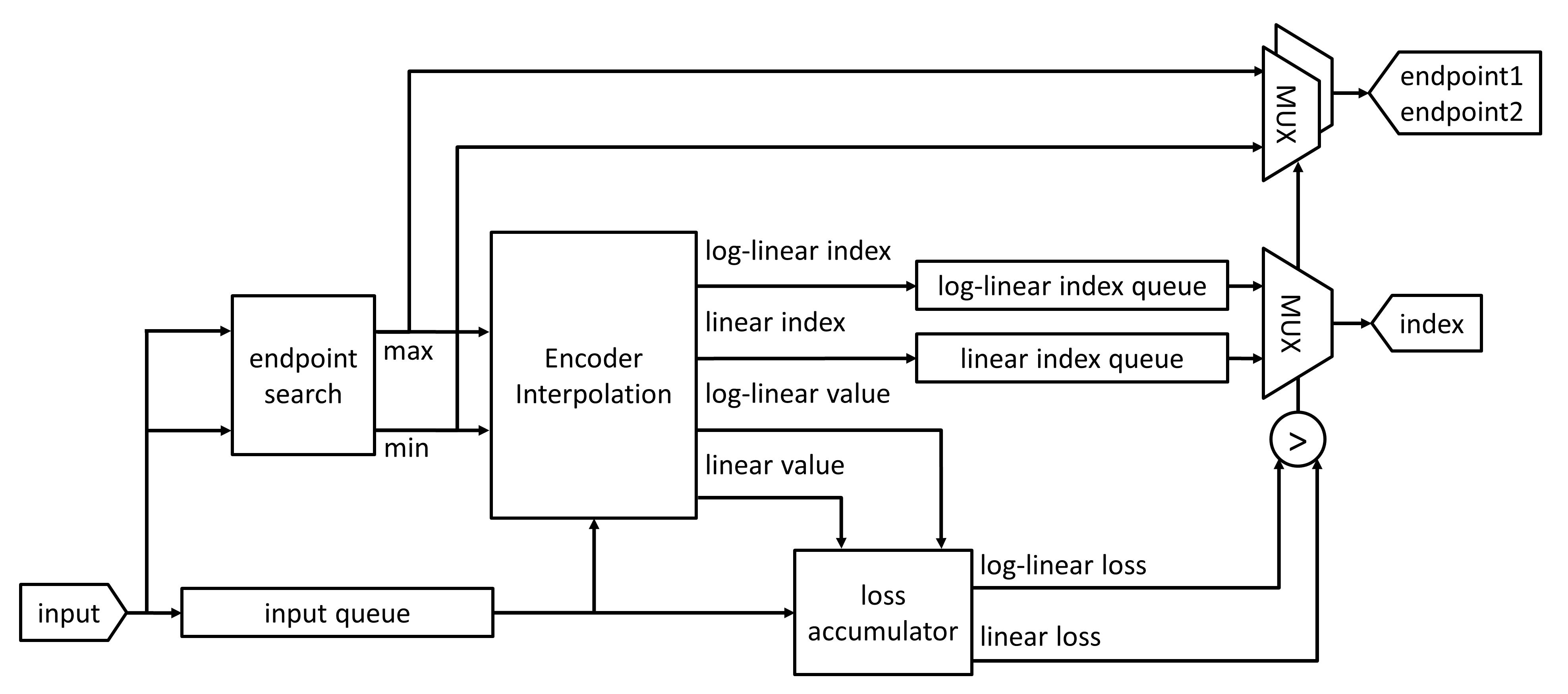}
\caption{Block diagram of ASC-CBR Encoder}
\label{hardware_encoder}
\end{figure}

Fig.~\ref{hardware_encoder} provides a block diagram of the proposed ASC-CBR encoder. The following paragraphs discuss the unscaled version, which processes the inputs sequentially. A comprehensive discussion on the hardware scaling procedure will be covered in a later section.

As depicted in Fig.~\ref{hardware_encoder}, for a specified input block, the initial step involves determining its endpoints using the endpoint search module. This module discerns the maximum and minimum endpoints within the block. Given that this procedure requires scanning all values within a block to establish the endpoints, an input queue is used to retain the previous input values. Armed with these endpoints and the input values, the encoder interpolation module discerns the nearest indexes and their associated values, utilizing both the revised linear scale and the log-linear scale. The choice between scales depends on the L1 losses associated with each.
During the loss computation process, the two types of index derived from the interpolation are also stored for later output. The loss accumulator module is responsible for calculating the L1 losses between the input values and the values obtained from interpolation.
In the final step, the encoder emits the indexes from the scale with a lower L1 loss, along with the endpoints. The output order of endpoint1 and endpoint2 is changed to signify the interpolation scale.

\begin{figure}[htbp]
    \centering
    \begin{subfigure}[h]{0.37\columnwidth}
        \centering
        \includegraphics[height=!,width=1.0\linewidth,keepaspectratio=true]{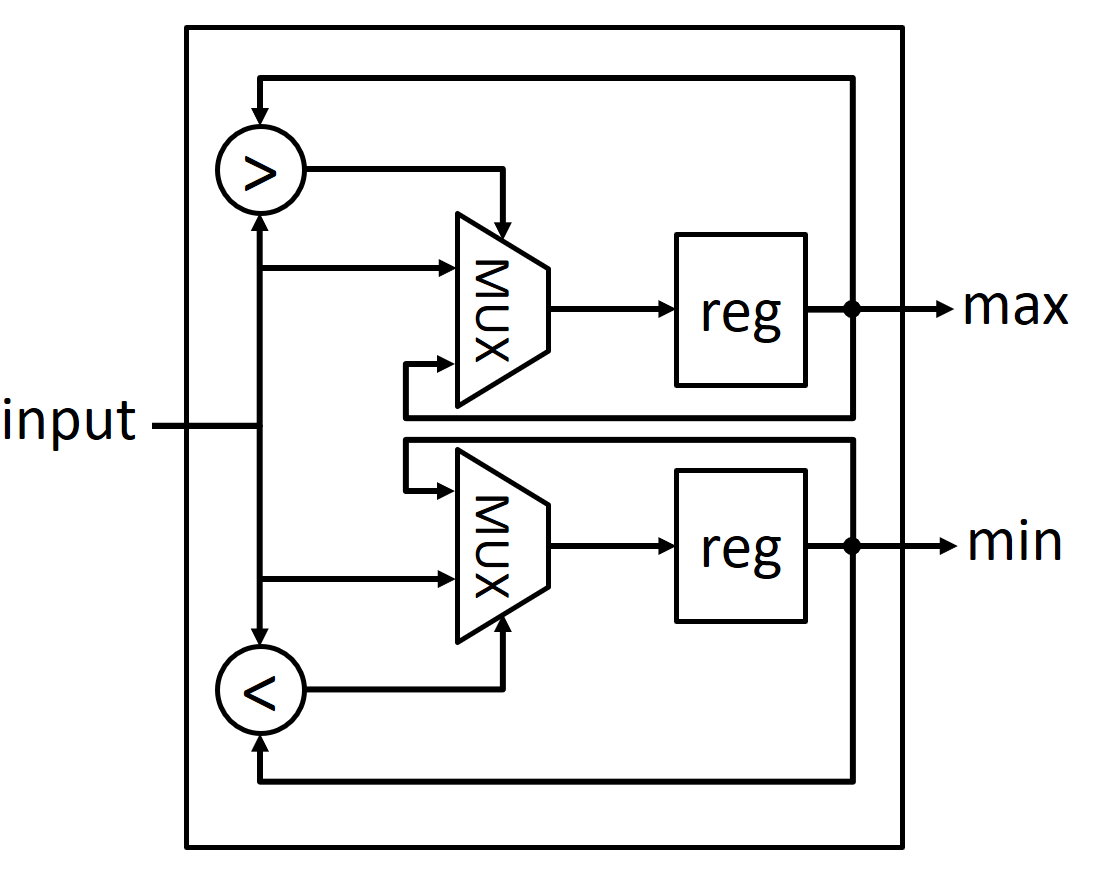}
        \caption{Endpoint search}
    \end{subfigure}
    \begin{subfigure}[h]{0.58\columnwidth}
        \centering
        \includegraphics[height=!,width=1.0\linewidth,keepaspectratio=true]{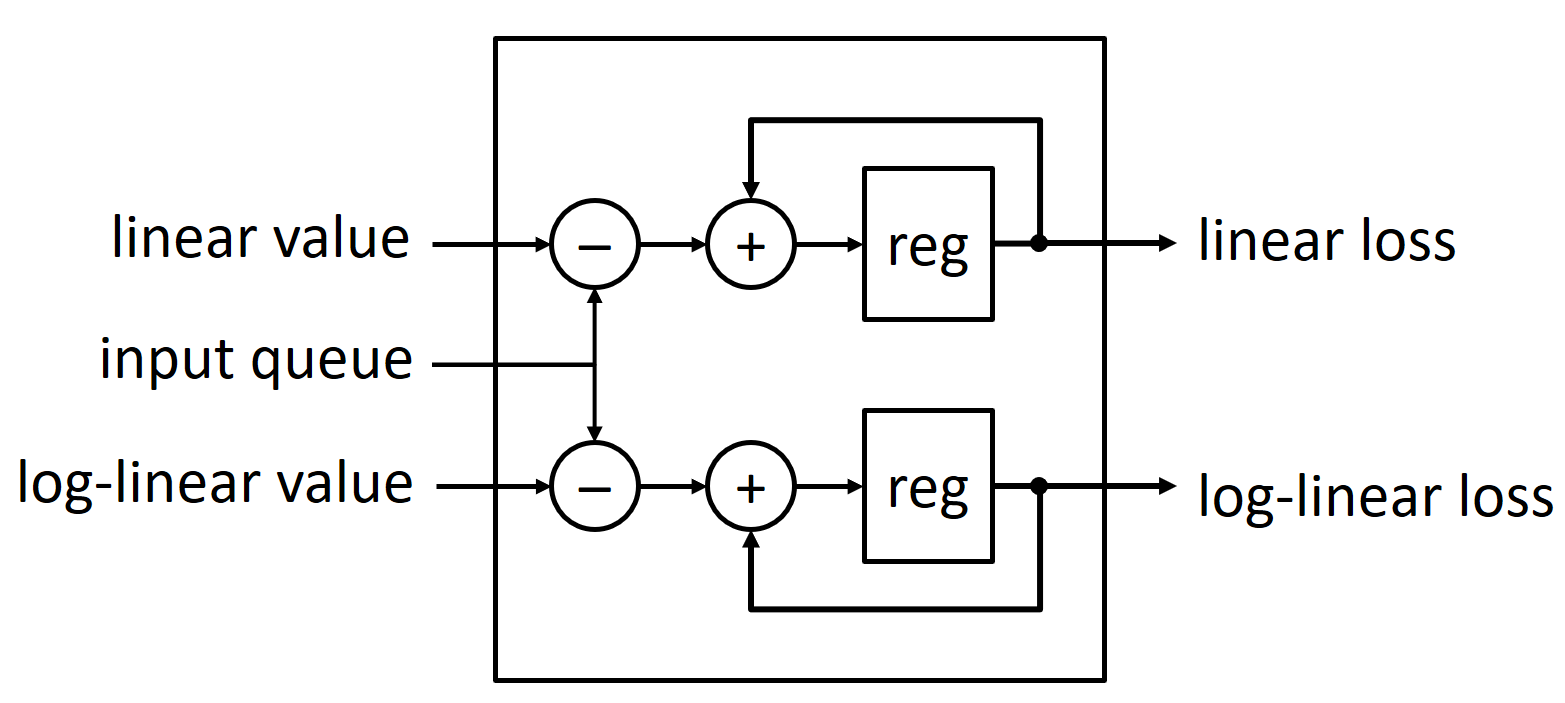}
        \caption{Loss accumulator}
    \end{subfigure}
    \caption{The detailed block diagram of (a) endpoint search, and (b)loss accumulator}
    \label{hardware_endpoint_loss}
\end{figure}

Fig.~\ref{hardware_endpoint_loss} illustrates the hardware components for endpoint search and loss accumulation, which are relatively simple. In the endpoint search module, two registers store the maximum and minimum values, updating with each input value. Meanwhile, the loss accumulator module computes the absolute differences between the input and interpolated values. Summing these differences in the loss registers yields the final L1 losses.

\begin{figure}[htbp]
\centering
\includegraphics[height=!,width=0.9\linewidth,keepaspectratio=true]{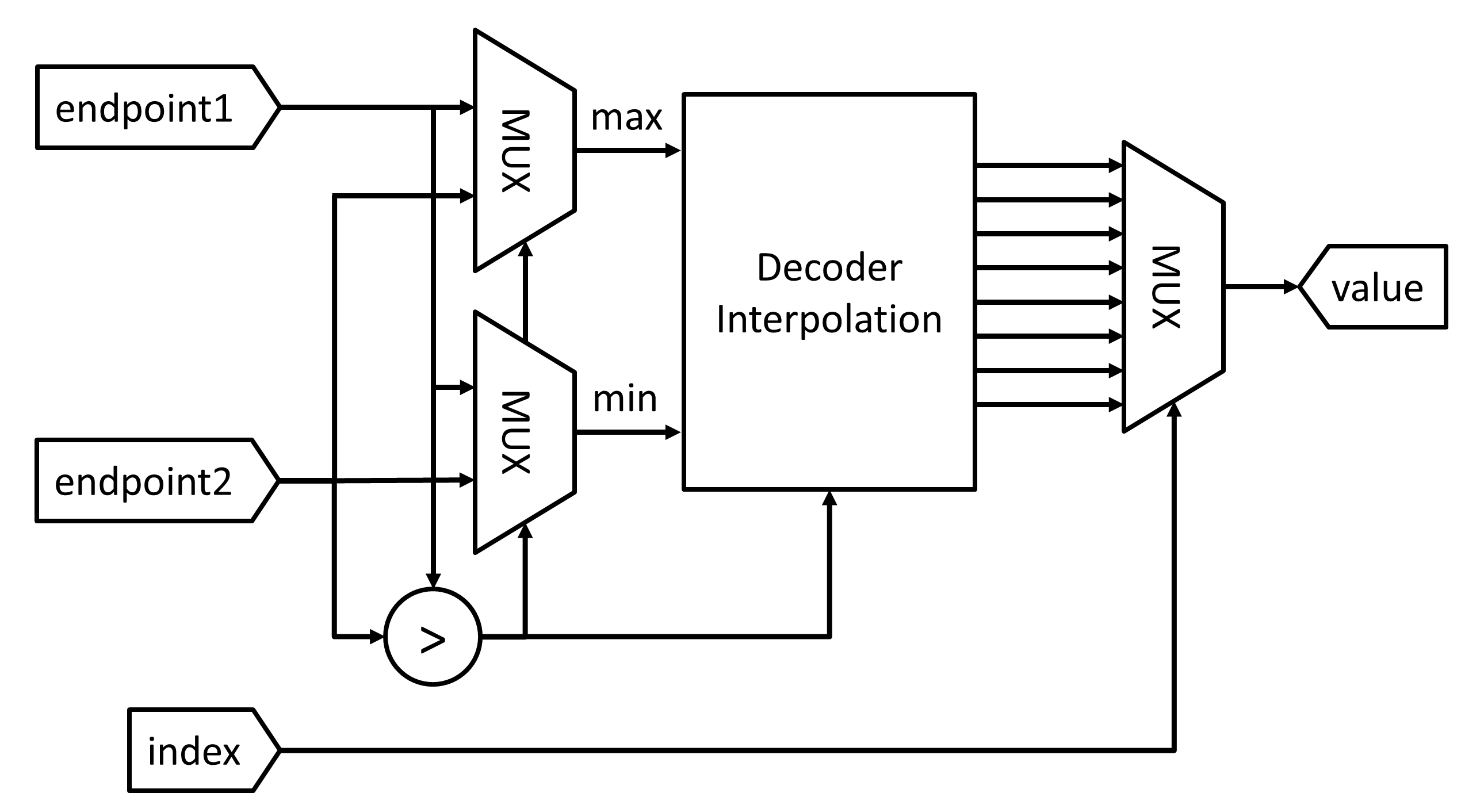}
\caption{Block diagram of ASC Decoder}
\label{hardware_decoder}
\end{figure}

The block diagram of the decoder, as shown in Fig.~\ref{hardware_decoder}, is notably more straightforward than its encoder counterpart. Initially, the two received endpoints are compared to ascertain the maximum and minimum endpoints for the 8-point interpolation across both scales. The appropriate scale is then chosen based on this comparison. Ultimately, using the input index, a multiplexer selects the output value from among the eight interpolation points.

\subsection{Interpolation}
In a straightforward implementation, determining the closest interpolation point for a given input entails calculating the distances between the input and the interpolation points. The input is then assigned to the interpolation point with the shortest distance. This approach requires a significant number of adders, multiplexers, and comparators, leading to prohibitive hardware costs.

\begin{figure}[htbp]
\centering
\includegraphics[height=!,width=1.0\linewidth,keepaspectratio=true]{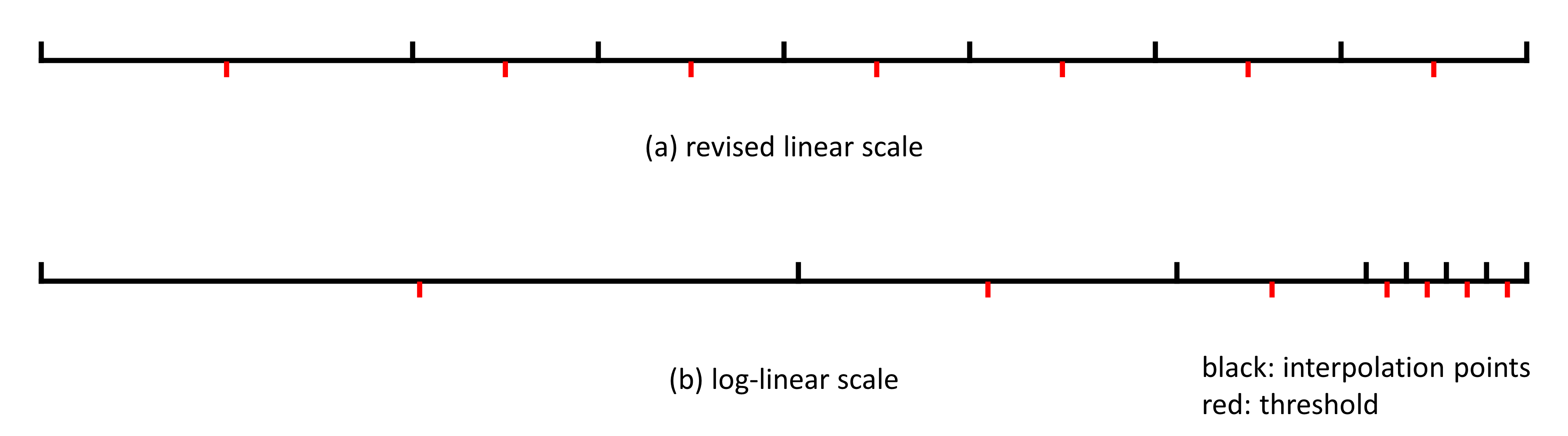}
\caption{The points and threshold values of (a) revised linear scale, (b) log-linear scale}
\label{hardware_interpolation_method}
\end{figure}

Fig.~\ref{hardware_interpolation_method} shows the two scales employed for interpolation: the revised linear scale and the log-linear scale. The two endpoints, $v_0$ and $v_7$, are sourced from the endpoint search module. The six intermediate points, ranging from $v_1$ to $v_6$, are determined based on the formulas presented in a subsequent paragraph.

Contrary to the straightforward approach, we incorporate the threshold concept to pinpoint the closest interpolation point. The seven threshold values, highlighted in red in the figure, are essentially the averages of adjacent points. When comparing a given input to all thresholds, if the input value exceeds $th_i$, it is allocated to $v_i$. If it exceeds several thresholds, the highest valid point is chosen. In contrast, if the input value falls below all thresholds, it is allocated to $v_0$.

This method presents multiple advantages. First, it sidesteps the intensive hardware component usage inherent in the straightforward approach. Second, the formulas for these thresholds, to be detailed in a subsequent paragraph, bear a striking resemblance to the formulas for interpolation points, facilitating hardware sharing. Lastly, since the thresholds remain constant within a block, scaling the hardware upwards incurs no additional costs.

\subsubsection{Formulas for interpolation points}
The formulas for the interpolation points are detailed in Table~\ref{table_interpolation_formula1}. This table features three columns representing the linear scale, the revised linear scale, and the log-linear scale. Within the table, \textit{M} denotes the maximum endpoint, while \textit{m} signifies the minimum endpoint. In the original linear scale, the interpolation points share a common denominator of 7, necessitating the use of dividers. These dividers are known for their complexity and challenges in hardware implementation. While it is possible to substitute the constant dividers with multipliers and shifters, the resulting hardware remains intricate. To address this, we modify the linear scale to adopt a common denominator of 8. As division by a power of two can be executed via bit-shifting, this modification obviates the need for dividers, leading to a marked reduction in hardware costs. Furthermore, it is noteworthy that v1, v2, and v4 in the revised linear scale align perfectly with v4, v5, and v6 in the log-linear scale. These pairs of points, highlighted in matching colors, present an avenue for hardware sharing.

\begin{table}[htbp]
\centering
\caption{The formulas for interpolation points}
\begin{tabular}{|c|c|c|c|}
\hline
\textbf{index}   & \textbf{linear scale} & \textbf{revised linear scale} & \textbf{log-linear scale} \\ \hline
$7$ & $M$                   & $M$                                     & $M$                                    \\ \hline
$6$ & $(m+6M)/7$   & $(2m+6M)/8$                    & $\textcolor{purple}{(m+M)/2}$ \\ \hline
$5$ & $(2m+5M)/7$  & $(3m+5M)/8$                    & $\textcolor{teal}{(3m+M)/4}$  \\ \hline
$4$ & $(3m+4M)/7$  & $\textcolor{purple}{(4m+4M)/8}$& $\textcolor{cyan}{(7m+M)/8}$  \\ \hline
$3$ & $(4m+3M)/7$  & $(5m+3M)/8$                    & $(29m+3M)/32$                 \\ \hline
$2$ & $(5m+2M)/7$  & $\textcolor{teal}{(6m+2M)/8}$  & $(15m+M)/16$                  \\ \hline
$1$ & $(6m+M)/7$   & $\textcolor{cyan}{(7m+M)/8}$   & $(31m+M)/32$                  \\ \hline
$0$ & $m$                   & $m$                                     & $m$      \\ \hline                           
\end{tabular}
\label{table_interpolation_formula1}
\end{table}

\begin{table}[htbp]
\centering
\caption{The formulas for interpolation points (shifted scale)}
\begin{tabular}{|c|c|c|c|}
\hline
\textbf{index} & \textbf{revised linear scale (shifted)} & \textbf{log-linear scale (shifted)} \\ \hline
$7$ & $\textcolor{purple}{(M-m)}$             & $\textcolor{purple}{(M-m)}$              \\ \hline
$6$ & $\textcolor{teal}{6(M-m)/8}$   & $\textcolor{purple}{(M-m)/2}$   \\ \hline
$5$ & $\textcolor{cyan}{5(M-m)/8}$   & $\textcolor{purple}{(M-m)/4}$   \\ \hline
$4$ & $\textcolor{purple}{4(M-m)/8}$ & $\textcolor{purple}{(M-m)/8}$   \\ \hline
$3$ & $\textcolor{teal}{3(M-m)/8}$   & $\textcolor{teal}{3(M-m)/32}$   \\ \hline
$2$ & $\textcolor{purple}{2(M-m)/8}$ & $\textcolor{purple}{(M-m)/16}$  \\ \hline
$1$ & $\textcolor{purple}{(M-m)/8}$  & $\textcolor{purple}{(M-m)/32}$  \\ \hline
$0$ & $0$                                     & $0$                \\ \hline
\end{tabular}
\label{table_interpolation_formula2}
\end{table}

\begin{table}[htbp]
\centering
\caption{The formulas for thresholds (shifted scale)}
\begin{tabular}{|c|c|c|c|}
\hline
\textbf{threshold} & \textbf{revised linear scale (shifted)} & \textbf{log-linear scale (shifted)} \\ \hline
$7$ & $\textcolor{violet}{7(M-m)/8}$   & $\textcolor{teal}{3(M-m)/4}$    \\ \hline
$6$ & $11(M-m)/16$                     & $\textcolor{teal}{3(M-m)/8}$    \\ \hline
$5$ & $9(M-m)/16$                      & $\textcolor{teal}{3(M-m)/16}$   \\ \hline
$4$ & $\textcolor{violet}{7(M-m)/16}$  & $\textcolor{violet}{7(M-m)/64}$ \\ \hline
$3$ & $\textcolor{cyan}{5(M-m)/16}$    & $\textcolor{cyan}{5(M-m)/64}$   \\ \hline
$2$ & $\textcolor{teal}{3(M-m)/16}$    & $\textcolor{teal}{3(M-m)/64}$   \\ \hline
$1$ & $\textcolor{purple}{(M-m)/16}$   & $\textcolor{purple}{(M-m)/64}$   \\ \hline         
\end{tabular}
\label{table_interpolation_formula3}
\end{table}

\subsubsection{Formulas for thresholds (shifted scale)}
Even with the adjustment mentioned above, the computation of interpolation points within both the revised linear scale and the log-linear scale remains reliant on a multitude of multipliers and adders. This setup provides only limited opportunities for hardware sharing. To further enhance this, we introduce the concept of scale shifting, as detailed below.

During interpolation within the hardware module, both scales are shifted by the value of the minimum endpoint. The shifted versions of these scales are delineated in Table~\ref{table_interpolation_formula2} for the interpolation points and in Table~\ref{table_interpolation_formula3} for the thresholds. In these tables, values that can be shared in hardware are highlighted using consistent colors.

The introduction of scale shifting yields two primary advantages. First, it eliminates the numerous multipliers and adders required for numerator calculations, thereby streamlining the hardware. Second, both the interpolation points and the thresholds of the shifted scales display pronounced similarities, which translates to a marked enhancement in hardware sharing.

Table~\ref{hardware_cost_comparison} details the hardware costs associated with computing interpolation points and thresholds. Clearly, the modification of the linear scale not only eliminates the need for dividers, but also marginally reduces hardware utilization. Crucially, the incorporation of scale shifting leads to a significant decrease in hardware costs, thanks to the extensive sharing of hardware resources.

\begin{table}[htbp]
\centering
\caption{Hardware cost for computing interpolation points and thresholds}
\label{hardware_cost_comparison}
\begin{tabular}{c|c|c|c|c|c}
\hline
                                                                          & \begin{tabular}[c]{@{}c@{}}Revised\\ linear\end{tabular} & \begin{tabular}[c]{@{}c@{}}Scale \\ Shifting\end{tabular} & Divider & Multiplier & Adder \\ \hline
\multirow{3}{*}{\begin{tabular}[c]{@{}c@{}}Hardware \\ Cost\end{tabular}} & –                                                        & –                                                         & 12      & 20         & 26    \\ \cline{2-6} 
                                                                          & v                                                        & –                                                         & 0       & 18         & 19    \\ \cline{2-6} 
                                                                          & v                                                        & v                                                         & 0       & 5          & 2     \\ \hline

\end{tabular}
\end{table}

\begin{figure}[htbp]
\centering
\includegraphics[height=!,width=1.0\linewidth,keepaspectratio=true]{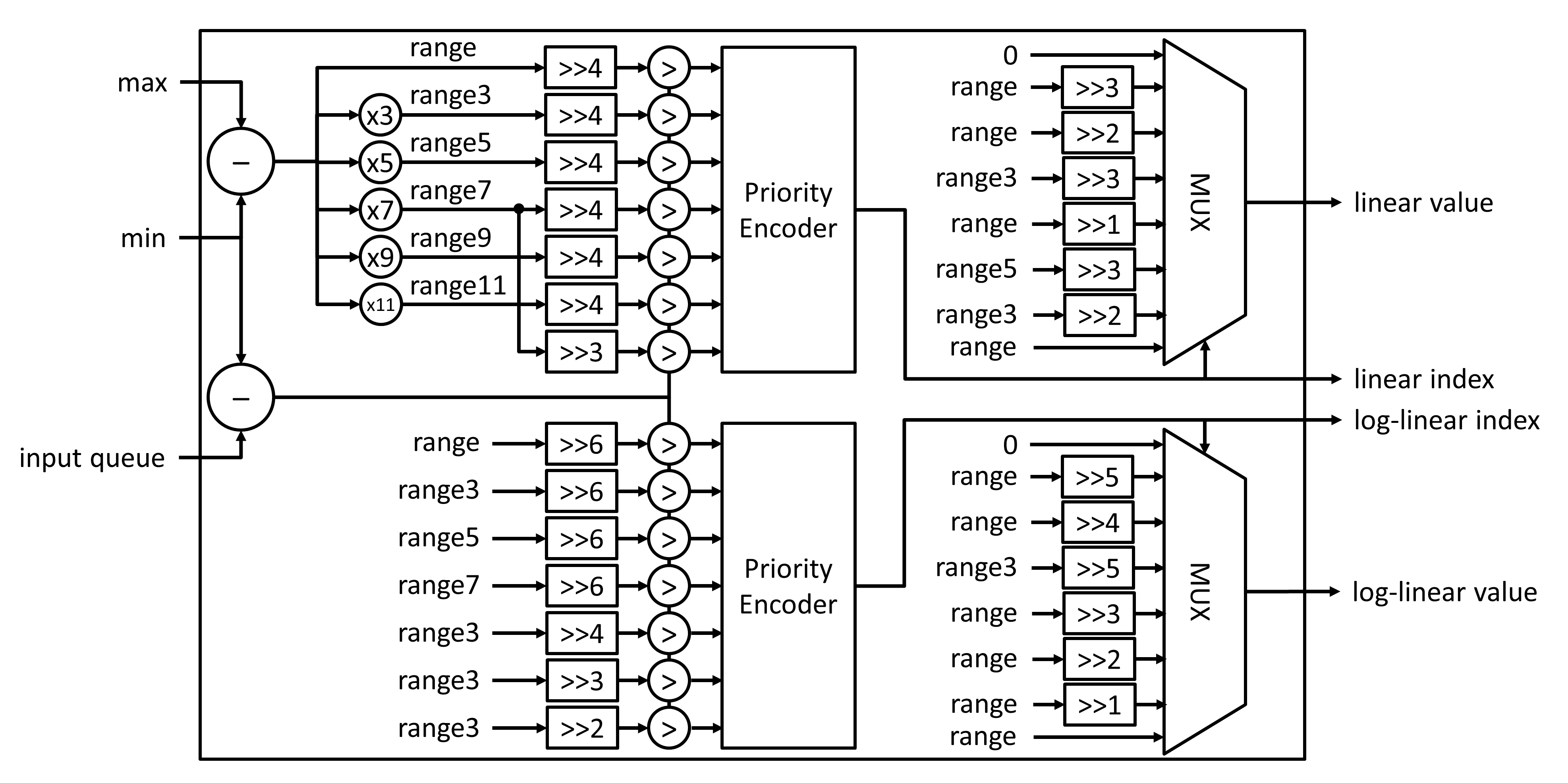}
\caption{The detailed block diagram of the encoder interpolation module}
\label{hardware_interpolation_encode}
\end{figure}

\subsubsection{Interpolation hardware}
The block diagram of the encoder interpolation module is depicted in Fig.~\ref{hardware_interpolation_encode}. We define the "range" as the difference between the maximum and minimum endpoints. By multiplying this range by various constants, we derive five multiples of the range. Remarkably, all interpolation points and thresholds are derived solely from these six variables, demonstrating significant hardware sharing. The input, sourced from the queue, undergoes a shift operation using the minimum value and is then compared against all thresholds. The priority encoder then processes the comparison results to produce the index. With this index in hand, the multiplexer selects an interpolation point to output.

\begin{figure}[htbp]
\centering
\includegraphics[height=!,width=0.9\linewidth,keepaspectratio=true]{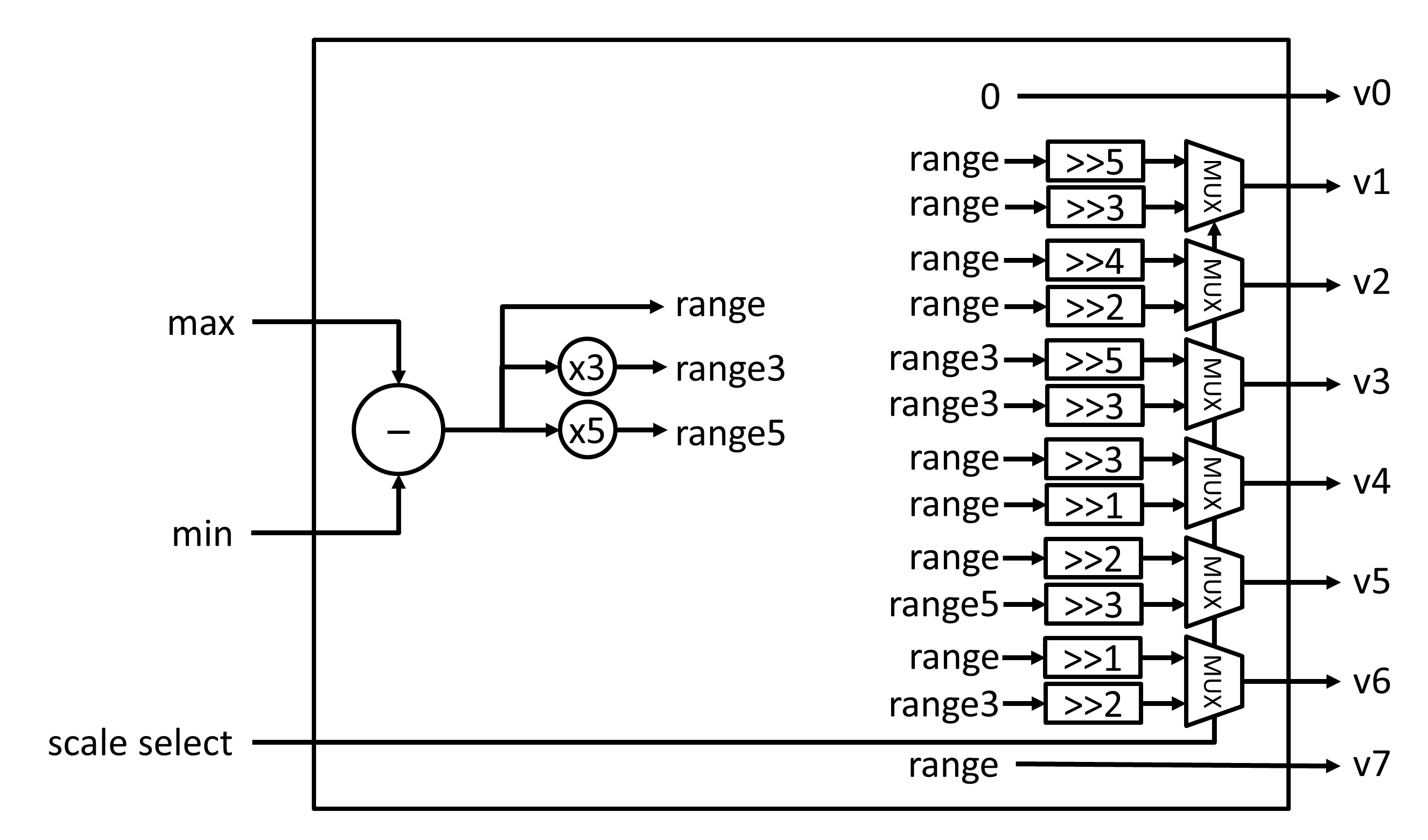}
\caption{The detailed block diagram of the decoder interpolation module}
\label{hardware_interpolation_decode}
\end{figure}

In contrast, Fig.~\ref{hardware_interpolation_decode} presents the decoder interpolation module. The computation of the range, its five multiples, and the interpolation points mirror the process described in the encoder interpolation module. However, in contrast to the encoder module, which outputs one of the eight interpolation points for both scales, the decoder module employs multiplexers to select one of the scales, subsequently outputting its eight interpolation points. This design choice excels in terms of hardware scalability. When scaling up the hardware, the interpolation points remain consistent for a block, which implies no additional cost for the decoder interpolation hardware. On the other hand, if we were to adopt the scheme from the encoder interpolation module, replication of the two multiplexers would be necessary, leading to increased hardware expenses.

\subsection{Hardware Scaling for Higher Throughput}

In this section, we discuss the procedure for scaling up the hardware for higher throughput. Our aim is to satisfy diverse throughput requirements up to the theoretical bandwidth of DDR5-6400. Using an 8x throughput scaling as an illustrative example, we will dissect the scaling processes for both the encoder and decoder.

Scaling up the encoder includes three main parts. First, in the endpoint search module, we need a multiplexer tree to find the maximum and minimum. Using the example of 8x scaling, we need a three-level multiplexer tree. Similarly, in the loss accumulator module, we use a three-level adder tree to sum the losses. We pipeline these two trees for shorter critical path. Second, the three queues must be reshaped simultaneously by widening and shortening the queues eight times. After the transformation, the queues still store all the values in a block, while doing so eight elements at a time. Lastly, in the encoder interpolation hardware, we need to make eight copies of the priority encoders and multiplexers, with each copy tackling one input.

As for the decoder, scaling it up only requires replicating the one multiplexer at the end of the block diagram. In this example, we copy this multiplexer eight times.

The proposed architecture is highly scalable. It performs simple operations and shares hardware components extensively. As a result, the cost of improving throughput scales sublinearly, which is shown in the implementation results.

\begin{figure*}[htbp]
\subfloat[AlexNet]{\includegraphics[trim={0cm 0 0cm 0}, clip, height=!,width=0.5\linewidth,keepaspectratio=true]{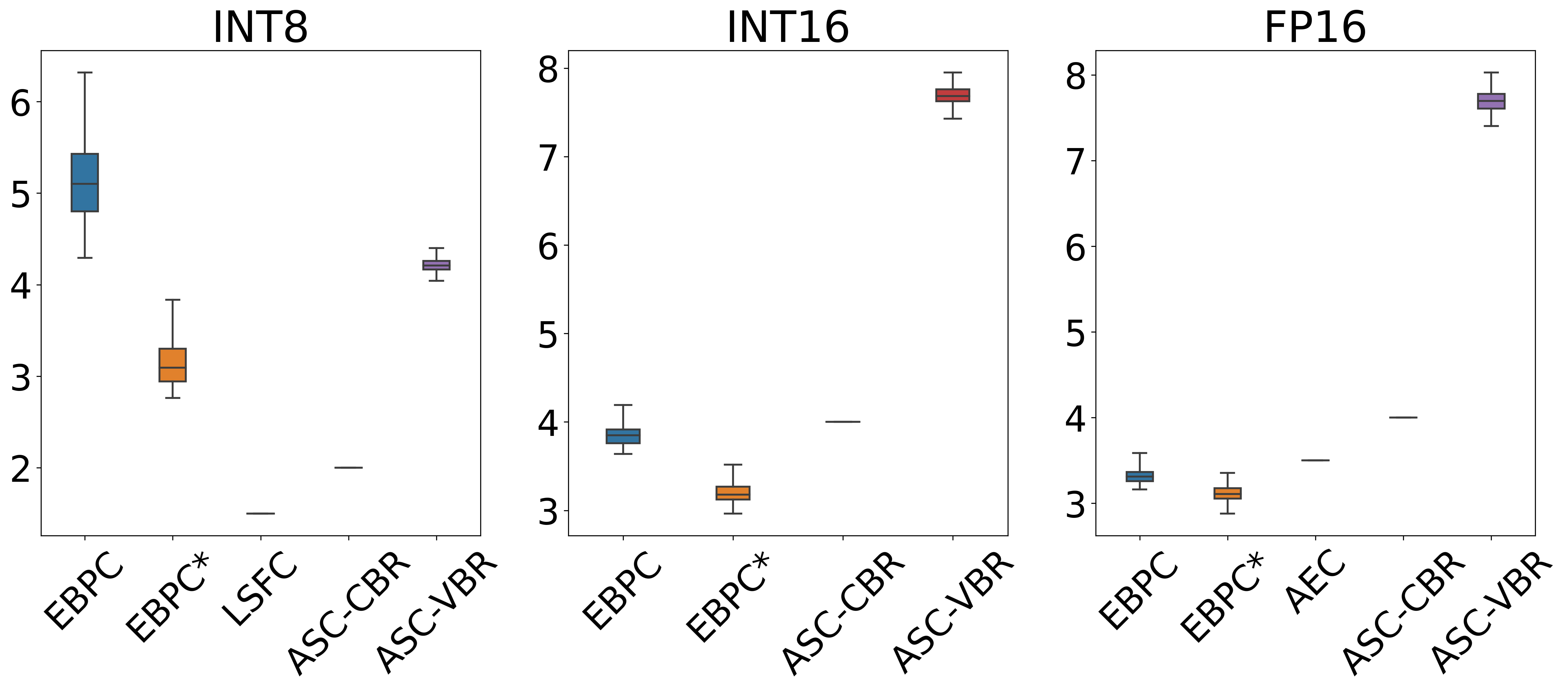}}
\subfloat[VGG16]{\includegraphics[trim={0cm 0 0cm 0}, clip, height=!,width=0.5\linewidth,keepaspectratio=true]{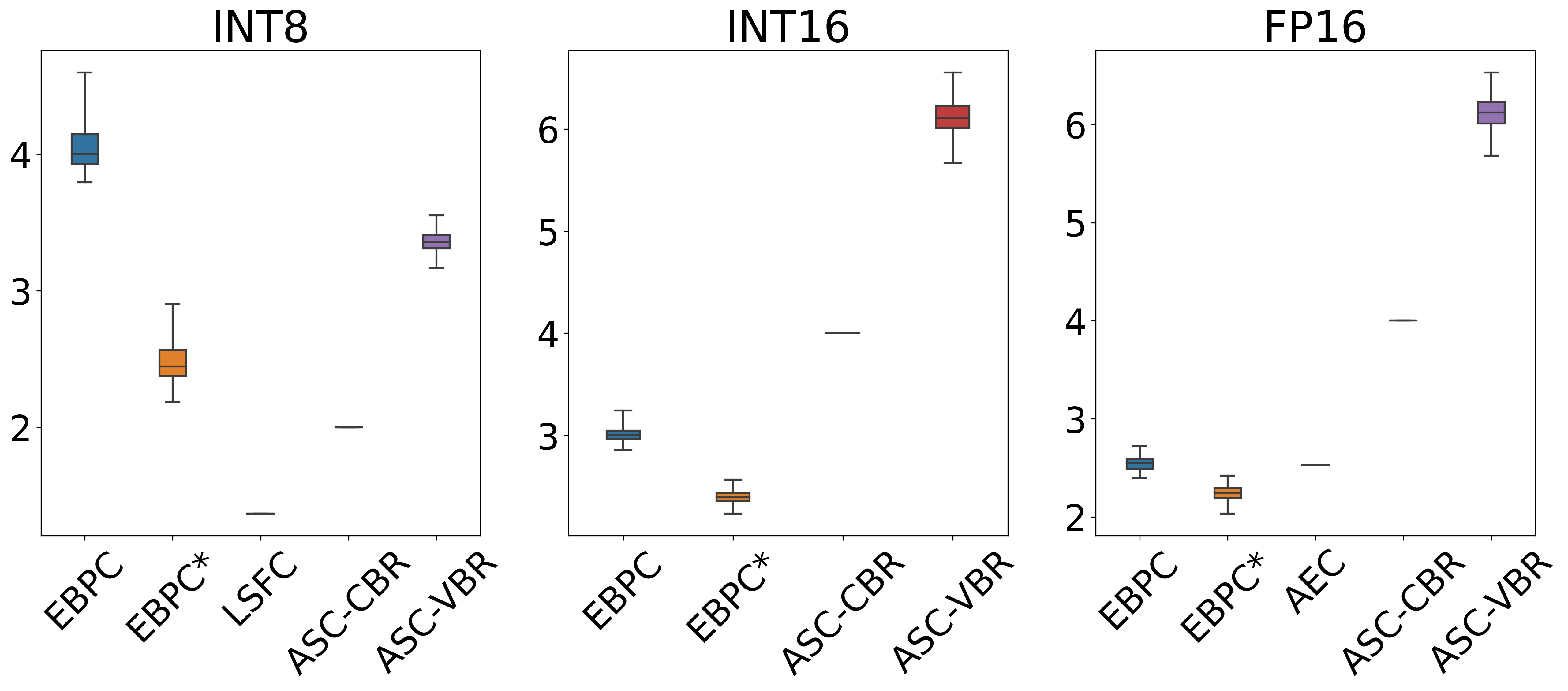}}
\newline
\subfloat[ResNet34]{\includegraphics[trim={0cm 0 0cm 0}, height=!,width=0.5\linewidth,keepaspectratio=true]{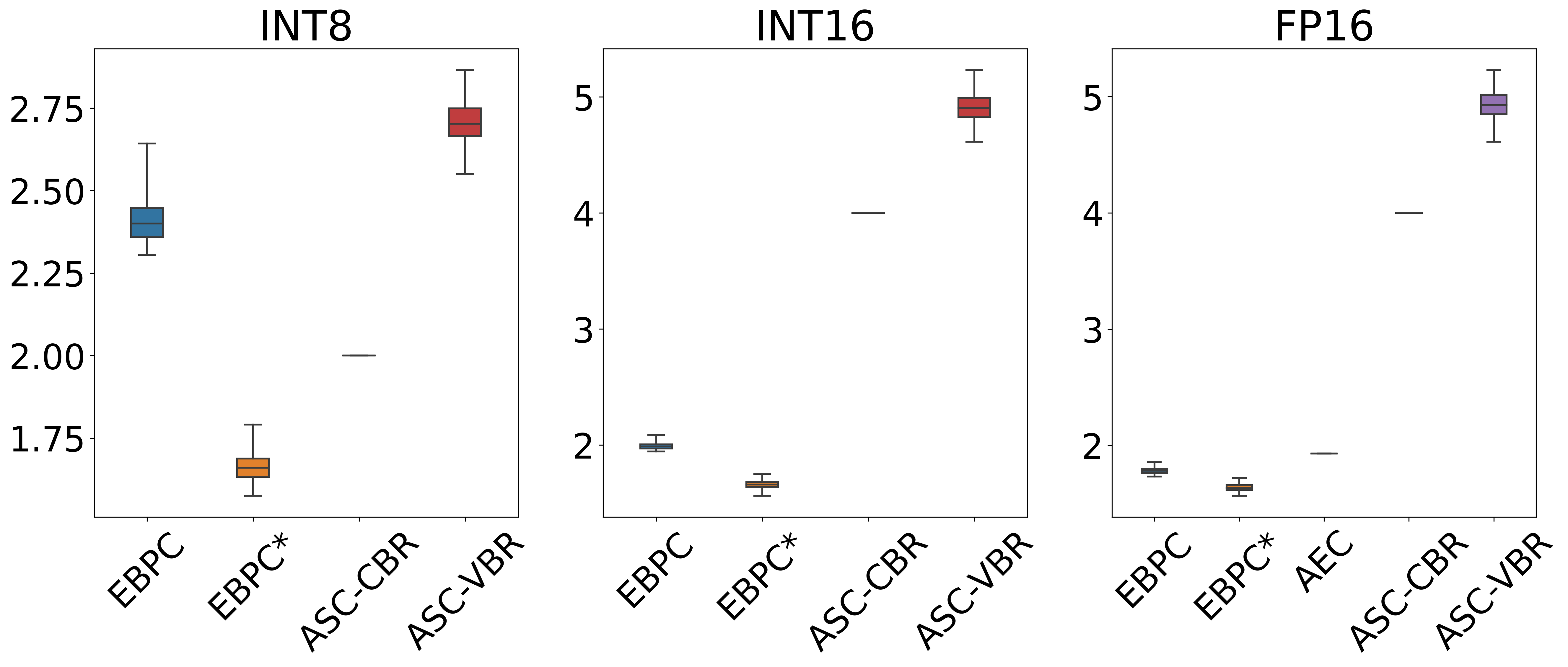}}
\subfloat[ViT-small]{\includegraphics[trim={0cm 0 0cm 0}, height=!,width=0.5\linewidth,keepaspectratio=true]{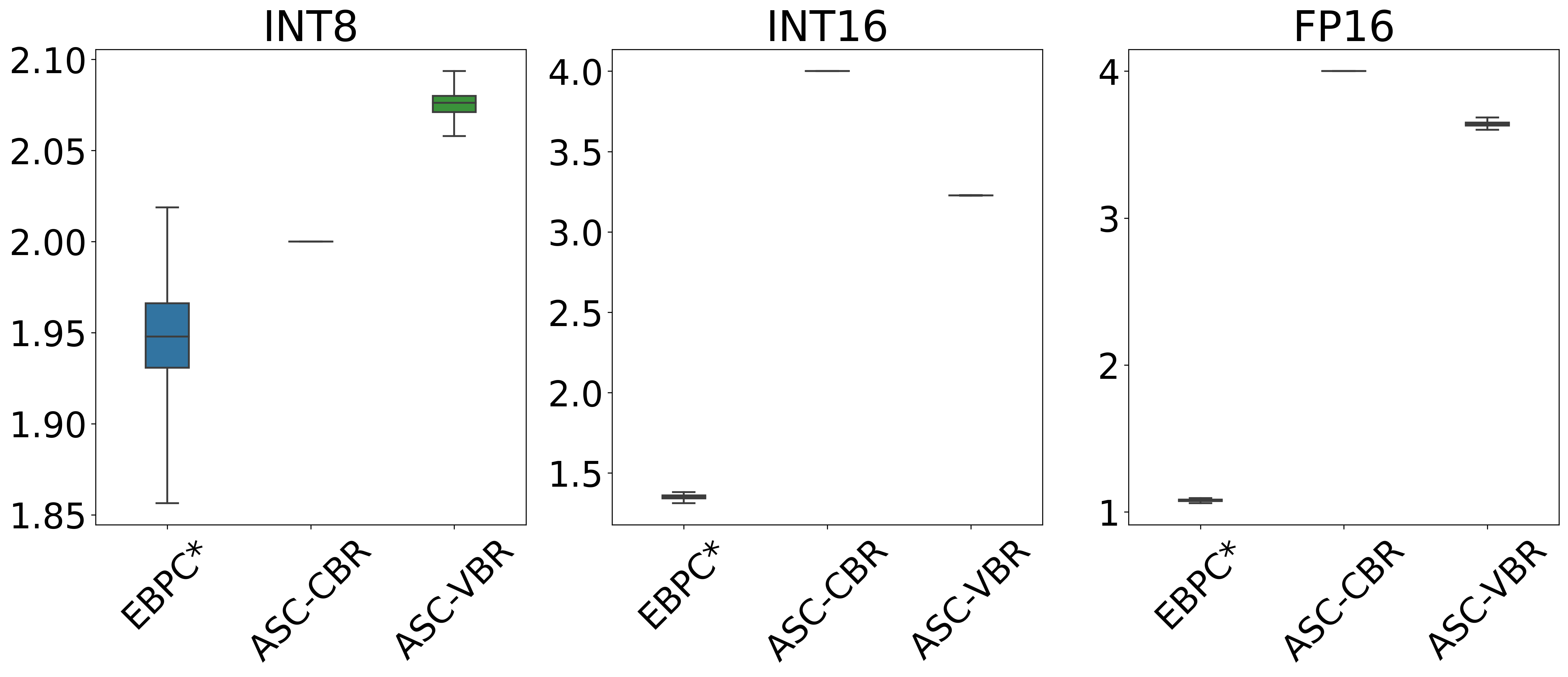}}

    \caption{Compression rates of various models and compression methods}
    \label{result_classification}
\end{figure*}

\begin{table}[htbp]
\centering
\caption{Accuracy drop (\%) of classification listed in the order of ASC-CBR/ASC-VBR}
\begin{tabular}{lccc}
\toprule
\diagbox{Model}{Format} & INT8 & INT16 & FP16 \\
\midrule
AlexNet & 0 / 0 & 0 / -0.18 & -0.03 / -0.19 \\
VGG16 & 0 / -0.24 & -0.08 / -0.25 & -0.03 / -0.15 \\
ResNet34 & -0.84 / -1.21 & -1.00 / -1.13 & -1.06 / -1.29 \\
ViT-small & -0.33 / -0.08 & -0.59 / -0.95 & -0.90 / -0.92 \\
\bottomrule
\end{tabular}
\label{result_classification_drop}
\end{table}

\begin{figure*}[htbp]
    \centering
    \includegraphics[height=!,width=0.8\linewidth,keepaspectratio=true]{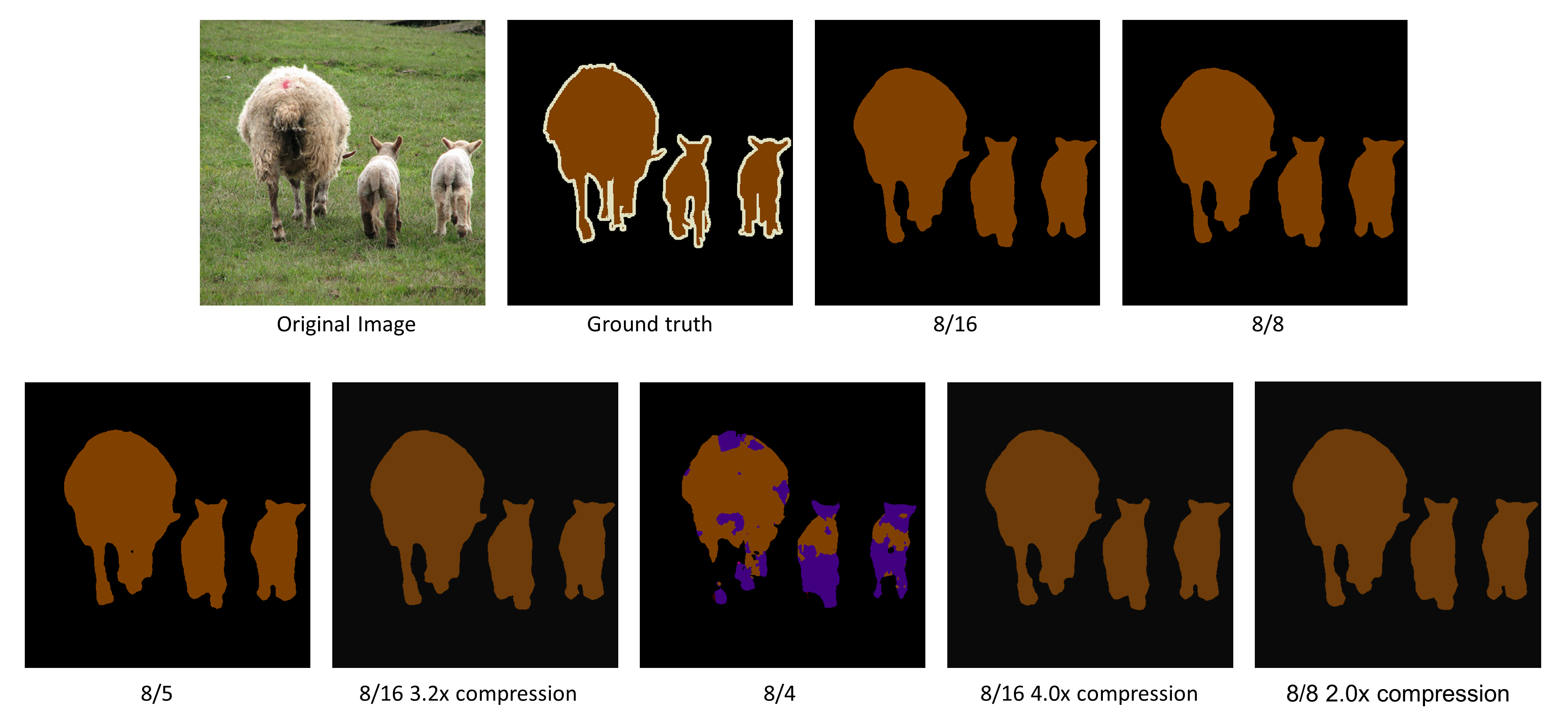}
    \caption{Qualitative results of segmentation for the PASCAL VOC dataset.}
    \label{result_segmentation}
\end{figure*}
\begin{figure*}[htbp]
    \centering
    \includegraphics[height=!,width=0.8\linewidth,keepaspectratio=true]{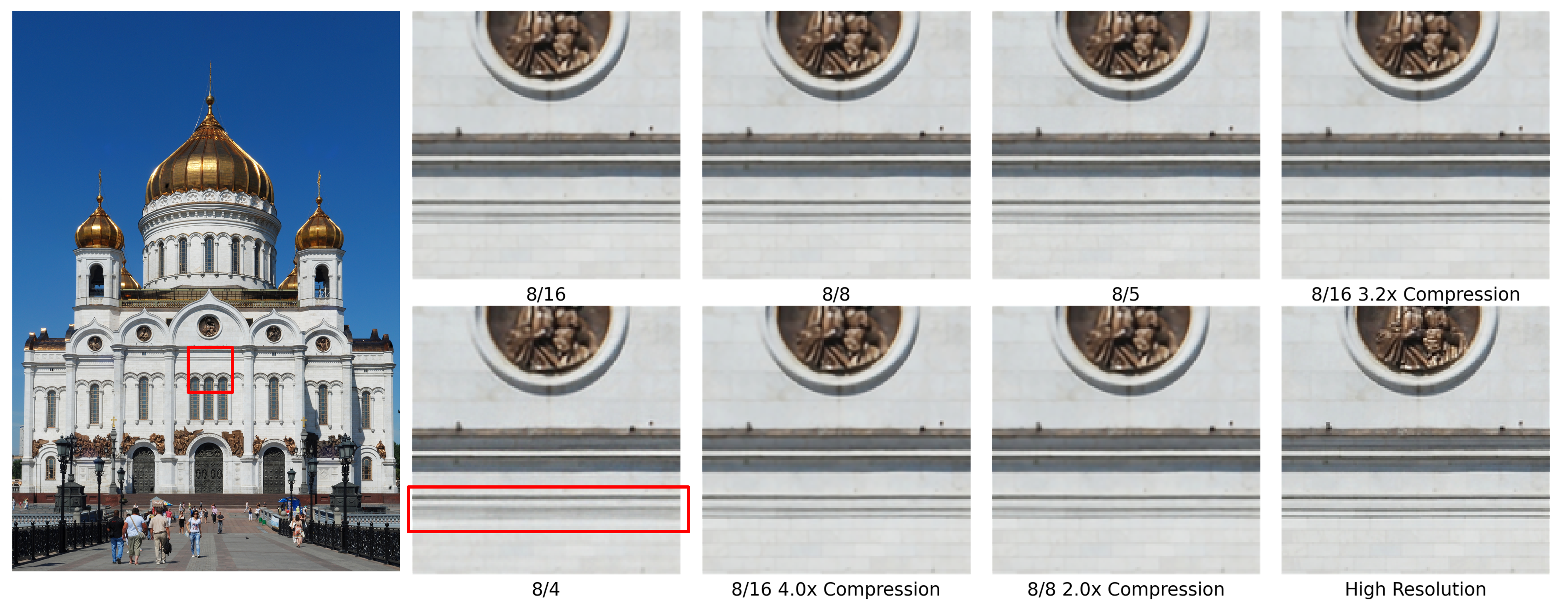}
    \caption{Qualitative results of super-resolution for the DIV2K dataset.}
    \label{result_superresolution}
\end{figure*}

\section{Experimental and Implementation Results}

\subsection{Experimental Setup}
The proposed algorithm has been evaluated in three tasks: classification, semantic segmentation, and super-resolution. To implement our approach, we used PyTorch/Torchvision and publicly available implementations of the selected networks as our baseline models with post-training static quantization. The experiments were carried out on an NVIDIA 24GB RTX 3090 GPU. Further details of our experimental setup are provided below.

The compressed bitstream is as follows: [\textit{endpoint1, endpoint2 (optional for one-point mode), index for each value of a block in a raster scan order}]. The block size and selection of one-point/two-point modes are fixed for a model. The scale will be the refined linear scale if endpoint1 is smaller than or equal to endpoint2. Otherwise, it will be the log-linear scale. The bit number for endpoints depends on the input data format. The index bit number will be 3 for eight interpolation points. All simulation results in this paper use the interpolation formula shown in Table~\ref{table_interpolation_formula2}. In subsequent experiments, the compression rate of our algorithm is defined as follows:

\begin{equation}
    \begin{split}
        compression\ rate = \frac{uncompressed\ bits}{compressed\ bits} \\
        =\frac{block_{size}\times Bit_{feature}}{Bit_{endpoint}\times NUM_{endpoint}+block_{size}\times 3}
    \end{split}
\end{equation}

For example, consider a block$_{size}$ of 16, the block shape is (2, 2, 4) according to \textit{cubical-like block shape}. Compression is performed in the two-endpoint mode for feature maps in the INT8 format. In this scenario, the compression rate is: (16 $\times$ 8) / ( 2 $\times$ 8 + 16 $\times$ 3 ) = 128 / 64 = 2.

The block$_{size}$ can be adjusted to achieve a trade-off between compression rate and performance. Increasing the block$_{size}$ leads to higher compression rates, but also results in more severe performance degradation due to compressing a larger number of values together. On the contrary, decreasing the block$_{size}$ yields the opposite effect, with lower compression rates and potentially improved performance.

\subsection{Ablation study of the proposed algorithm}
This subsection shows the simulation results for different parameters used in the proposed algorithm.

\subsubsection{Channel Indexing and Reordering}
Table~\ref{table_indexing_reordering} presents the results of various channel indexing and \textit{similarity-based reordering} methods. The table indicates that the heuristic method experiences less performance degradation due to its superior channel ordering. In terms of channel indexing, even when \textit{similarity-based reordering} is applied and a smaller block$_{size}$ of 8 is used, there is a notable performance degradation with the shared channel indexing method. This underscores the limited correlation between channels within the feature maps, rendering shared channel indexing less suitable for feature map compression. On the other hand, when \textit{independent channel indexing} is employed, as illustrated in the table's last row, there is no observable performance degradation. This highlights the efficacy of \textit{independent channel indexing} in compressing feature maps without compromising performance.

Consequently, in our subsequent experiments, we will employ \textit{independent channel indexing}. Furthermore, when \textit{similarity-based reordering} is implemented, we will opt for the heuristic method to reorder the channels.

\begin{table}[htbp]
\centering
\caption{Result of channel indexing methods and \textit{similarity-based reordering} on VGG16.}
\begin{tabular}{c|c|c|c|c}
\hline
\textbf{\begin{tabular}[c]{@{}c@{}}Channel  \\  Indexing\end{tabular}} &
  \textbf{\begin{tabular}[c]{@{}c@{}}(endpoint,\\ block$_{size}$)\end{tabular}} &
  \textbf{\begin{tabular}[c]{@{}c@{}}Compr.\\ Rate\end{tabular}} &
  \textbf{\begin{tabular}[c]{@{}c@{}}Reordering\\ \end{tabular}} &
  \textbf{\begin{tabular}[c]{@{}c@{}}Accuracy\\ Drop (\%)\end{tabular}} \\ \hline
\multirow{6}{*}{Shared} & \multirow{3}{*}{(1,   32)} & \multirow{3}{*}{3.934} & –         & -23.48 \\
                        &                            &                        & greedy    & -13.13 \\
                        &                            &                        & heuristic & -11.01 \\ \cline{2-5} 
                        & \multirow{3}{*}{(1,   8)}  & \multirow{3}{*}{2.285} & –         & -1.95  \\
                        &                            &                        & greedy    & -1.22  \\
                        &                            &                        & heuristic & -1.17  \\ \hline
Independent             & (1,   8)                   & 2.0                    & –         & 0.00   \\ \hline
\end{tabular}
\label{table_indexing_reordering}
\end{table}

\subsubsection{Cubical-Like Block Shape}
\begin{figure}[htbp]
\centering
\includegraphics[height=!,width=1.0\linewidth,keepaspectratio=true]{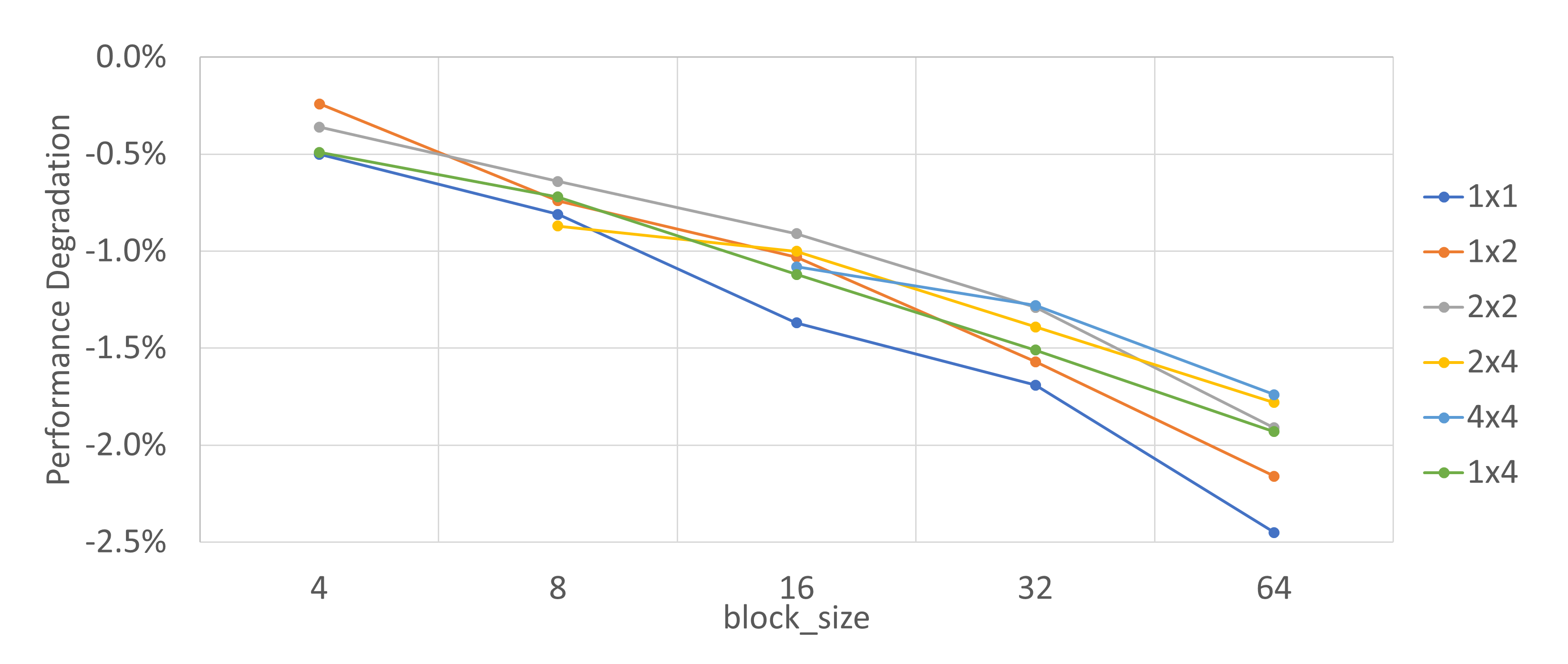}
\caption{Performance degradation with different block shapes and block sizes for ResNet34.}
\label{result_cubical-liked_block_shape}
\end{figure}

Fig.\ref{result_cubical-liked_block_shape} illustrates the performance implications of varying block shapes and block sizes. It should be noted that while the same compression rate is achieved for a specific block$_{size}$, the choice of block shape profoundly influences performance. This underscores the importance of selecting an optimal block shape. To determine the block shape, we opt for the most cubical-like one. For example, considering a block$_{size}$ of 1024, the step-by-step calculation to derive the most cubical-like block shape proceeds as follows: (1, 1, 1024), (2, 2, 256), (4, 4, 64), and (8, 8, 16). We halt the procedure once we recognize that 16 is 2x of 8. As a result, the final selection is (8, 8, 16). This choice aligns closely with the best-performing block shape as depicted in Table\ref{table_cubical-liked_block_shape}. However, it is worth noting that the block shapes (1, 2, 2) and (2, 2, 1) are equally cubical in nature, differing only in orientation. Collectively, the findings confirm that our proposed \textit{cubical-like block shape} consistently delivers optimal performance across various scenarios.

\begin{table}[htbp]
\centering
\caption{The best accuracy block shape and proposed \textit{cubical-like block shape} for different block$_{size}$}
\begin{tabular}{c|c|c|c|c|c}
\hline
\textbf{block$_{size}$}               & 4           & 8           & 16          & 32          & 64          \\ \hline
\textbf{best}  & (1,   2, 2) & (2,   2, 2) & (2,   2, 4) & (4,   4, 2) & (4,   4, 4) \\ \hline
\textbf{proposed}  & (2,   2, 1) & (2,   2, 2) & (2,   2, 4) & (4,   4, 2) & (4,   4, 4) \\ \hline
\end{tabular}%
\label{table_cubical-liked_block_shape}
\end{table}

\subsubsection{Switchable Endpoint Mode}
\begin{table}[htbp]
\centering
\caption{Compression performance for various models with different endpoint modes}
\begin{tabular}{c|c|c|c|c}
\hline
\textbf{Model} &
  \textbf{\begin{tabular}[c]{@{}c@{}}Sparsity\\(\%) \end{tabular}} &
  \textbf{\begin{tabular}[c]{@{}c@{}}Compression\\ Rate\end{tabular}} &
  \textbf{\begin{tabular}[c]{@{}c@{}}(endpoint,\\ block$_{size}$)\end{tabular}} &
  \textbf{\begin{tabular}[c]{@{}c@{}}Accuracy\\ Drop(\%)\end{tabular}} \\ \hline
\multirow{2}{*}{AlexNet}   & \multirow{2}{*}{74.48} & \multirow{10}{*}{2.0} & (1,   8) & +0.07 \\
                           &                          &                       & (2, 16)  & -0.03 \\ \cline{1-2} \cline{4-5} 
\multirow{2}{*}{VGG16}     & \multirow{2}{*}{60.43} &                       & (1,   8) & +0.06 \\
                           &                          &                       & (2, 16)  & +0.03 \\ \cline{1-2} \cline{4-5} 
\multirow{2}{*}{ResNet34}  & \multirow{2}{*}{43.82} &                       & (1,   8) & -0.84 \\
                           &                          &                       & (2, 16)  & -0.98 \\ \cline{1-2} \cline{4-5} 
\multirow{2}{*}{ViT-small} & \multirow{2}{*}{35.04} &                       & (1,   8) & -0.33 \\
                           &                          &                       & (2, 16)  & -0.74 \\ \cline{1-2} \cline{4-5} 
\multirow{2}{*}{\begin{tabular}[c]{@{}c@{}}DeepLabv3+\\ (ResNet50)\end{tabular}} &
  \multirow{2}{*}{77.19} &
   &
  (1,   8) &
  -1.25 \\
                           &                          &                       & (2, 16)  & -1.68\\ \hline
\end{tabular}
\label{table_sparsity_one_endpoint}
\end{table}

Table~\ref{table_sparsity_one_endpoint} showcases the compression performance across different models. When comparing the two configurations, the one-endpoint mode demonstrates a smaller performance drop than the two-endpoint mode at equivalent compression rates. This indicates that when a model has a notably high sparsity level, presuming the minimum value in a block to be zero results in negligible performance impact.

To explain the usage of the two-endpoint mode, we conducted experiments on the EDSR model for the super-resolution task. The residual block in EDSR consists of a convolution-ReLU-convolution branch, where the sparsity is 91.29\%, 11.47\% and 51.38\% for the first convolution after ReLU, the second convolution, and average, respectively. Table~\ref{table_sparsity_two_endpoint} presents the compression performance of EDSR, where the baseline PSNR is 35.0487. In this case, the two-endpoint mode significantly outperforms the one-endpoint mode at the same compression rate. This finding highlights the importance of considering the specific characteristics of the model when selecting the endpoint mode. Despite EDSR's overall high sparsity, compressing convolution layers with exceptionally low sparsity using the one-endpoint mode leads to marked performance degradation. Therefore, it is imperative to select the compression mode that aligns with the model's sparsity distribution.

\begin{table}[htbp]
\centering
\caption{Compression performance for EDSR with different endpoint modes}
\begin{tabular}{c|c|c}
\hline
\textbf{Compression   Rate} & \textbf{(endpoint, block$_{size}$)} & \textbf{PSNR} \\ \hline
\multirow{2}{*}{2.0}        & (1,   8)                         & 32.3142     \\
                            & (2,   16)                        & 35.0018     \\ \hline
\end{tabular}%
\label{table_sparsity_two_endpoint}
\end{table}

\subsubsection{Adaptive Scale Interpolation}
\begin{table}[htbp]
\centering
\caption{Compression performance of ViT-small using different interpolation methods}
\begin{tabular}{c|c|c|c}
\hline

\begin{tabular}[c]{@{}c@{}}Compression\\  Rate\end{tabular} & \begin{tabular}[c]{@{}c@{}}(endpoint, \\ block$_{size}$)\end{tabular} & \begin{tabular}[c]{@{}c@{}}Interpolation\\  Method\end{tabular} & \begin{tabular}[c]{@{}c@{}}Accuracy \\ Drop\end{tabular} \\ \hline

\multirow{2}{*}{2.0}   & \multirow{2}{*}{(1,   8)}  & Linear           & -0.72\% \\
                       &                            & Adaptive & -0.45\% \\ \hline
\multirow{2}{*}{2.285} & \multirow{2}{*}{(1,   16)} & Linear           & -2.03\% \\
                       &                            & Adaptive  & -0.85\% \\ \hline
\multirow{2}{*}{2.461} & \multirow{2}{*}{(1,   32)} & Linear           & -6.00\% \\
                       &                            & Adaptive & -1.42\% \\ \hline
\end{tabular}%
\label{table_adaptive_scale}
\end{table}

Table~\ref{table_adaptive_scale} displays the compression performance of ViT-small using various interpolation techniques. The \textit{adaptive scale interpolation} consistently outperforms the linear scale interpolation, with the performance gap increasing as the block$_{size}$ expands. A larger block$_{size}$ accentuates the spatial and channel-wise distance between the most distant value pairs, reducing intra-block correlation and heightening the likelihood of outliers. By skillfully handling pronounced outliers and maintaining accuracy for smaller values, the \textit{adaptive scale interpolation} establishes itself as the superior method.

\subsection{Task-specific evaluation}
\subsubsection{Classification}
The classification task is evaluated in the ILSVRC2012 validation set with three pretrained CNN models (AlexNet, VGG16, and ResNet34) from PyTorch/Torchvision and one pretrained transformer model (ViT-small) from~\cite{timm}. We follow the data preprocessing guidelines recommended by those sources.

Fig.~\ref{result_classification} shows the compression rates of different models using various compression methods. Each box plot in the figure is based on 250 random samples from the ILSVRC2012 validation set. The results with the EBPC label are obtained from the original paper~\cite{EBPC}, while the results labeled as EBPC* are generated by us using their publicly available code. The compression rates presented in the EBPC* column are derived from the actual feature maps extracted from our models, ensuring a fairer comparison. In the original papers of LSFC~\cite{least-square-fitting} and AEC~\cite{Area_efficient}, only the average compression rates are provided. As a result, the columns for LSFC and AEC may appear to offer fixed compression rates, but they actually have variable compression rates. In ASC-CBR, the one-endpoint mode and a block$_{size}$ of 8 are used, except for FP16 ViT-small, which uses the two-endpoint mode and a block$_{size}$ of 16. In ASC-VBR, the two-endpoint mode and a block$_{size}$ of 32 are used for all models and formats.

The compression rates of EBPC* vary across different models and input samples, owing to variations in sparsity levels between models and inputs. On the contrary, our proposed ASC-CBR achieves a fixed compression rate across all models and input samples. Furthermore, ASC-CBR outperforms EBPC* in terms of compression rates, with the exception of AlexNet and VGG16, which exhibit high levels of sparsity. If a fixed compression rate and hardware scalability are not essential requirements, ASC-VBR offers an even higher compression rate, surpassing all the listed compression methods in all formats.

Another useful observation pertains to the compression rates achieved for the vision transformer. It is evident that methods relying on sparsity prove ineffective in achieving high compression rates for the vision transformer. This disparity can be attributed to the activation function employed by the vision transformer, namely Gaussian error linear units (GELU), which differs from the commonly used rectified linear unit (ReLU). Unlike ReLU, the GELU activation function changes inputs based on their percentile rather than gating them based on their sign. As a result, the sparsity level with the vision transformer is considerably lower, which poses challenges for methods that rely on sparsity exploitation.

Table~\ref{result_classification_drop} illustrates the impact on classification accuracy for different models using our proposed method. Both AlexNet and VGG16 exhibit nearly lossless performance, maintaining accuracy levels comparable to their uncompressed counterparts. On the other hand, ResNet34 and ViT-small experience a moderate decrease in accuracy, with a maximum drop of -1.29\%.

\subsubsection{Semantic Segmentation}
The semantic segmentation is evaluated the Pascal VOC dataset~\cite{PASCOLVOC} with the widely used DeepLabv3+~\cite{DeepLabV3Plus} with a ResNet50 backbone.

\begin{table}[htbp]
\centering
\caption{Segmentation results for PASCAL VOC dataset.}
\begin{tabular}{c|c|c|c|c|c}
\hline
\textbf{\begin{tabular}[c]{@{}c@{}}Effective\\ Bits\end{tabular}} &
  \textbf{Method} &
  \textbf{\begin{tabular}[c]{@{}c@{}}Quant\\ (W/A)\end{tabular}} &
  \textbf{\begin{tabular}[c]{@{}c@{}}endpoint,   \\ block$_{size}$\end{tabular}} &  
  \textbf{\begin{tabular}[c]{@{}c@{}}Compr. \\ Rate\end{tabular}} &
  \textbf{\begin{tabular}[c]{@{}c@{}}mIoU \\ Drop\end{tabular}} \\ \hline
\multirow{2}{*}{5} & (AQ)      & 8/5  & –        & –   & -5.84\%  \\
                   & (MQ,   C) & 8/16 & (1,   8) & 3.2 & -1.07\%  \\ \hline
\multirow{3}{*}{4} & (AQ)      & 8/4  & –        & –   & -36.59\% \\
                   & (MQ,   C) & 8/16 & (1, 16)  & 4.0 & -1.55\%  \\
                   & (MQ,   C) & 8/8  & (1, 8)   & 2.0 & -1.30\%  \\ \hline
\end{tabular}
\label{SS_table}
\end{table}

\begin{table}[htbp]
\centering
\caption{Super-resolution results for DIV2K dataset.}
\begin{tabular}{c|c|c|c|c|c}
\hline
\textbf{\begin{tabular}[c]{@{}c@{}}Effective\\ Bits\end{tabular}} &
  \textbf{Method} &
  \textbf{\begin{tabular}[c]{@{}c@{}}Quant.\\ (W/A)\end{tabular}} &
  \textbf{\begin{tabular}[c]{@{}c@{}}endpoint,   \\ block$_{size}$\end{tabular}} &  
  \textbf{\begin{tabular}[c]{@{}c@{}}Compr.   \\ Rate\end{tabular}} &
  \textbf{\begin{tabular}[c]{@{}c@{}}PSNR   \\ Drop\end{tabular}} \\ \hline
\multirow{2}{*}{5} & (AQ)      & 8/5  & –         & –   & -0.3137 \\
                   & (MQ,   C) & 8/16 & (2,   16) & 3.2 & -0.0364 \\ \hline
\multirow{3}{*}{4} & (AQ)      & 8/4  & –         & –   & -0.9200 \\
                   & (MQ,   C) & 8/16 & (2,   32) & 4.0 & -0.0400 \\
                   & (MQ,   C) & 8/8  & (2,   16) & 2.0 & -0.0563 \\ \hline
\end{tabular}
\label{SR_table}
\end{table}

\begin{table*}[h!]
\centering
\caption{Comparison with other feature map compression works, where "-" denotes data unavailable. Cls: classification, OD: object detection, Seg: semantic segmentation, and SR: superresolution.}
\label{hardware_comparison}
\begin{tabular}{|c|c|c|c|c|c|c|cc|}
\hline
Method                                                                                   & EBPC~\cite{EBPC} & AEC~\cite{Area_efficient}  & Wavelet~\cite{Wavelet-compression} & DCT~\cite{DCT-memoryefficient} & LSFC~\cite{least-square-fitting} & Real-Time~\cite{real-time-superresolution} & \multicolumn{2}{c|}{ASC}                \\ \hline
Performance                                                                             & Lossless       & Lossless      & Lossy             & Lossy         & Lossy          & Lossy               & \multicolumn{2}{c|}{Lossy}              \\ \hline
Task                                                                                    & Cls            & Cls           & OD, Seg, SR       & OD            & Cls            & SR                  & \multicolumn{2}{c|}{Cls, Seg, SR}       \\ \hline
Transform                                                                               & –              & –             & Haar wavelet      & DCT           & –              & –                   & \multicolumn{2}{c|}{–}                  \\ \hline
Compression rate                                                                        & Variable       & Variable      & Fixed             & Variable      & Variable       & Fixed               & \multicolumn{2}{c|}{Fixed/Variable}     \\ \hline
Supported format                                                                        & FP   / INT     & FP            & FP   / INT        & INT           & INT            & INT                 & \multicolumn{2}{c|}{FP   / INT}         \\ \hline
Technology (nm)                                                                         & 65             & 130           & –                 & 28            & 28             & FPGA                & \multicolumn{2}{c|}{28}                 \\ \hline
Clock   Rate (MHz)                                                                      & 600            & –             & –                 & 700           & 800            & 150                 & \multicolumn{2}{c|}{1600}               \\ \hline
Throughput (Moutputs/s)                                                                 & 480            & –             & –                 & –             & –              & –                   & \multicolumn{1}{c|}{1600}    & 51200    \\ \hline
Gate Count                                                                              & 5890           & 2085          & –                 & –             & –              & –                   & \multicolumn{1}{c|}{6135}    & 46956    \\ \hline
\begin{tabular}[c]{@{}c@{}}Area Efficiency\\  (Throughput/Gate count)\end{tabular} & 81.49k         & –             & –                 & –             & –              & –                   & \multicolumn{1}{c|}{260.79k} & 1090.38k \\ \hline
Power (mW)                                                                              & –              & –             & –                 & –             & –              & –                   & \multicolumn{1}{c|}{6.139}   & 37.845   \\ \hline
\end{tabular}
\end{table*}

\begin{table}[h!]
\centering
\caption{Gate count of our implementation.}
\label{APR_gate_count}
\begin{tabular}{cccc}
\hline
        & Scaling              & INT8  & INT16  \\ \hline
Encoder & \multirow{3}{*}{1$\times$}  & 6614  & 12963  \\
Decoder &                      & 1058  & 4233   \\
Total   &                      & 7672  & 17196  \\ \hline
Encoder & \multirow{3}{*}{32$\times$} & 44709 & 105820 \\
Decoder &                      & 6614  & 12963  \\
Total   &                      & 51323 & 118783 \\ \hline
\end{tabular}
\end{table}

\begin{table}[h!]
\centering
\caption{Power consumption of our implementation.}
\label{APR_power}
\begin{tabular}{cccc}
\hline
        & Scaling              & INT8 (mW)    & INT16 (mW)   \\ \hline
Encoder & \multirow{3}{*}{1$\times$}  & 7.972 & 8.299 \\
Decoder &                      & 0.457 & 0.887 \\
Total   &                      & 8.429 & 9.186 \\ \hline
Encoder & \multirow{3}{*}{32$\times$} & 31.3  & 62    \\
Decoder &                      & 10.9  & 18.7  \\
Total   &                      & 42.2  & 80.7  \\ \hline
\end{tabular}
\end{table}

Table~\ref{SS_table} presents the experimental results obtained from the PASCAL VOC dataset. To make a comparison, we introduce the concept of effective bits, which refers to the number of bits saved in memory per value of the feature map. In our analysis, we consider two approaches: aggressive quantization (AQ) and a combination of mild quantization (MQ) with our proposed adaptive scale feature map compression (C). In AQ, the model is directly quantized from FP32 to the effective bits, while in the other approach, the model is initially quantized to a higher bit precision before applying the adaptive scale feature map compression.

The table shows that AQ causes a significant drop in mIoU, while the combination of MQ with our proposed ASC method enables us to achieve storage reduction while preserving performance. Qualitative results are shown in Fig.~\ref{result_segmentation}. Consistent with the quantitative findings, our segmentation maps closely resemble their MQ baseline counterparts. Furthermore, compared to AQ results with the same effective bits, the output segmentation maps of ASC exhibit improved accuracy.

\subsubsection{Super Resolution}
We also performed a superresolution task in which the objective is to reconstruct a high-resolution image from a single low-resolution image. To address this task, we selected the popular EDSR network~\cite{EDSR} with its basic configuration for 2$\times$ scaling, trained on the DIV2K dataset~\cite{DIV2K}. Following a similar approach to~\cite{Wavelet-compression}, we focus our compression efforts solely on the body of the network. The model has a baseline PSNR of 35.0581.

Table~\ref{SR_table} presents the experimental results from the DIV2K dataset. Again, we observed that AQ leads to a significant reduction in PSNR, while combining MQ with ASC reduces storage while preserving performance. Fig.~\ref{result_superresolution} shows the qualitative results of the cropped regions obtained from different approaches. The use of AQ leads to the emergence of artifacts within the cropped regions, as highlighted. In contrast, the ASC result remains visually plausible and has minimal artifacts.

\begin{figure}[htbp]
\centering
\includegraphics[height=!,width=0.4\linewidth,keepaspectratio=true]{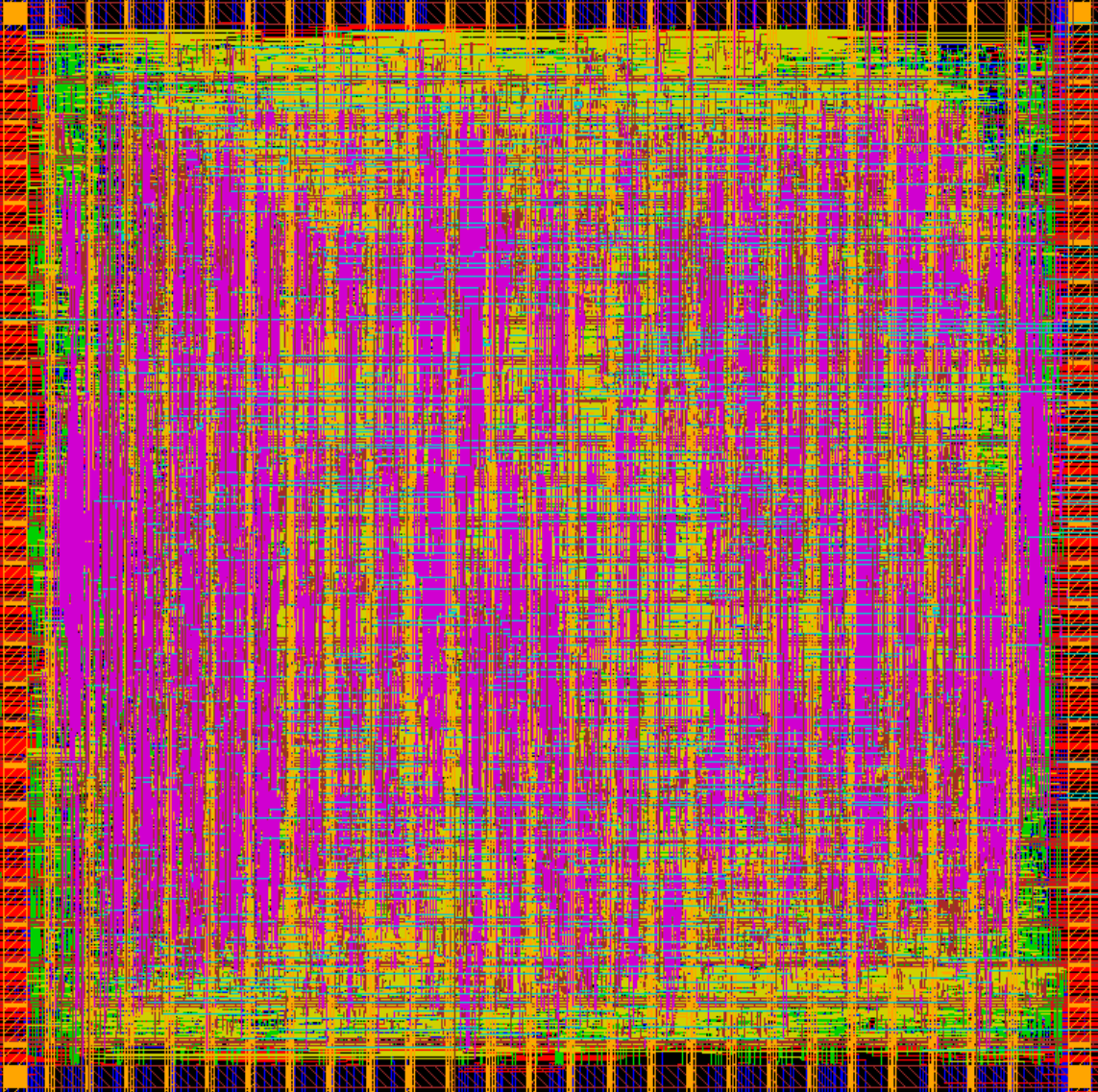}
\caption{Design layout of the 32$\times$ scaling.}
\label{APR}
\end{figure}

\subsection{Hardware implementation}
The design has been designed with Verilog, synthesized with Synopsys Design Compiler, placed and routed with Cadence Innovus using TSMC 28nm CMOS technology at 1.6GHz. Hardware power consumption is estimated by Synopsys PrimeTime PX. Fig.~\ref{APR} shows the layout. Table~\ref{APR_gate_count} and Table~\ref{APR_power} show the gate count and power analysis. As shown in the table, the gate count and the power of the decoder are much smaller than those of the encoder due to the simple decoder nature of a texture-based method. Furthermore, our proposed architecture is highly scalable, with a 32$\times$ increase in throughput requiring only 6.68$\times$ and 6.9$\times$ the hardware cost for INT8 and INT16 respectively. A similar increase can also be found for power consumption. This is due to our parallel algorithm and hardware optimization. In comparison, to increase 32$\times$ throughput for EBPC~\cite{EBPC} will need 32$\times$ area costs due to their serial processing nature. Furthermore, it is worth noting that with a clock frequency of 1.6GHz, the 32$\times$ hardware is able to reach our target, the theoretical bandwidth of DDR5-6400.

\subsection{Design comparison}
Table~\ref{hardware_comparison} shows a comparison to other feature map compression works. In this table, \cite{Wavelet-compression} proposed the software algorithm only without hardware implementation. \cite{EBPC, Area_efficient} did not report some implementation results. \cite{DCT-memoryefficient, least-square-fitting, real-time-superresolution} provide only the information for the entire DLA, with no result for a single encoder and decoder. Thus, the corresponding data are not available.

Compared to other designs, our work has the highest throughput with a comparable gate count. Moreover, due to the highly scalable architecture, it can be scaled up to process 51.2GB/s~\cite{DDR5}, which is the theoretical bandwidth of DDR5-6400. Importantly, the hardware cost increases sublinearly during scaling, enhancing power and area efficiency.

\section{Conclusion}

This paper introduces the adaptive scale feature map compression, ASC, a method that is simple, scalable, and effective in alleviating memory bandwidth and buffer size constraints in deep learning accelerators. ASC exploits feature map attributes to propose several techniques to enhance the compression rate with minimum accuracy loss, which is applicable to both fixed-point and floating-point formats.
The constant bitrate variant delivers a compression rate of 2.0$\times$ for 8-bit data and 4.0$\times$ for 16-bit data, maintaining near-lossless performance across a range of models and tasks. Moreover, the variable bitrate version surpasses existing benchmarks, achieving compression rates as high as 4.2$\times$ for 8-bit and 7.69$\times$ for 16-bit data.
From a hardware perspective, our design significantly reduces the area cost of interpolation by refined linear scale and shifted scale strategies. These not only simplify the design, but also promote hardware sharing. Implemented with TSMC 28nm technology, our system runs at 1.6GHz and just needs a modest gate count of 6135 for 8-bit data. The architecture demonstrates exceptional scalability: a 32-fold increase in throughput requires only a 7.65$\times$ increase in hardware resources. Given its potential to decrease memory bandwidth, power usage, and on-chip buffer size, we are confident that our methodology can greatly benefit resource-limited deep learning accelerators.

\bibliographystyle{IEEEtran}

\bibliography{IEEEabrv, bib/ieeeBSTcontrol,bib/thesis}

\begin{IEEEbiography}[{\includegraphics[width=1in,height=1.25in,clip,keepaspectratio]{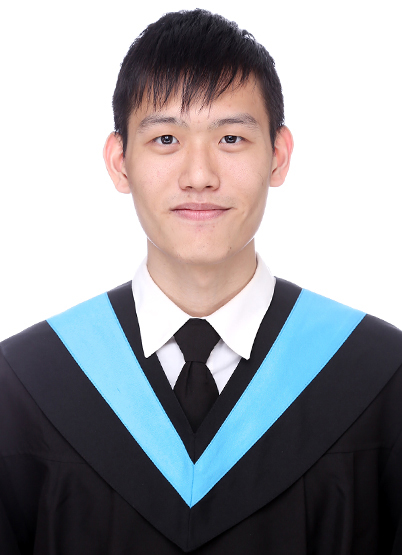}}]{Yuan Yao}
received received the B.S., and M.S. degrees in electronic engineering from the National Yang Ming Chiao Tung University, Hsinchu, Taiwan, in 2023. He is currently working at MediaTek, Hsinchu, Taiwan. His current research interests include deep learning accelerator design and VLSI. 

\end{IEEEbiography}

\begin{IEEEbiography}[{\includegraphics[width=1in,height=1.25in,clip,keepaspectratio]{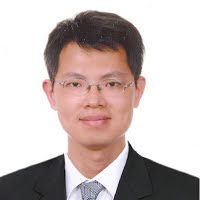}}]{Tian-Sheuan Chang}
	(S’93–M’06–SM’07)
	received the B.S., M.S., and Ph.D. degrees in electronic engineering from National Chiao-Tung University (NCTU), Hsinchu, Taiwan, in 1993, 1995, and 1999, respectively. 
	
	From 2000 to 2004, he was a Deputy Manager with Global Unichip Corporation, Hsinchu, Taiwan. In 2004, he joined the Department of Electronics Engineering, NCTU (as National Yang Ming Chiao Tung University (NYCU) in 2021), where he is currently a Professor. In 2009, he was a visiting scholar in IMEC, Belgium. His current research interests include system-on-a-chip design, VLSI signal processing, and computer architecture.
	
	Dr. Chang has received the Excellent Young Electrical Engineer from Chinese Institute of Electrical Engineering in 2007, and the Outstanding Young Scholar from Taiwan IC Design Society in 2010. He has been actively involved in many international conferences as an organizing committee or technical program committee member.
\end{IEEEbiography}
\end{document}